\documentclass{article}
\usepackage[final]{nejlt}
\usepackage{times}
\usepackage{latexsym}
\usepackage{graphicx}
\usepackage[noglossbreaks,owncounter]{covington}
\usepackage{url}
\usepackage{csquotes}
\usepackage{hyperref}
\usepackage{isomath}
\usepackage{multirow}
\usepackage{booktabs}
\usepackage{amsmath}
\usepackage{xcolor}
\usepackage{relsize}

\newcommand{\word}[1] {`\textit{#1}'}

\newcommand{\sense}[1] {{\smaller[0.5]`\textsc{{#1}}'}}

\title{Contextualized language models for semantic change detection: lessons learned}

\author{
Andrey Kutuzov, University of Oslo, Norway {\tt \small andreku@ifi.uio.no} \and
Erik Velldal, University of Oslo, Norway {\tt \small erikve@ifi.uio.no} \and
Lilja Øvrelid, University of Oslo, Norway {\tt \small liljao@ifi.uio.no}  
}

\begin{document}

\abstract{We present a qualitative analysis of the (potentially erroneous) outputs of contextualized embedding-based methods for detecting diachronic semantic change. First, we introduce an ensemble method outperforming previously described contextualized approaches. This method is used as a basis for an in-depth analysis of the degrees of semantic change  predicted for English words across 5 decades. Our findings show that contextualized methods can often predict high change scores for words which are not undergoing any real diachronic semantic shift in the lexicographic sense of the term (or at least the status of these shifts is questionable). Such challenging cases are discussed in detail with examples, and their linguistic categorization is proposed. Our conclusion is that pre-trained contextualized language models are prone to confound changes in lexicographic senses and changes in contextual variance, which naturally stem from their distributional nature, but is different from the types of issues observed in methods based on static embeddings. Additionally, they often merge together syntactic and semantic aspects of lexical entities. We propose a range of possible future solutions to these issues.}

\maketitle

\section{Introduction}

Lexical semantic change detection (LSCD) is a relatively recent sub-field within natural language processing. However, comprehensive surveys of data-driven  modeling of diachronic semantic change are already available \cite{tang_2018,kutuzov:survey,tahmasebi2018survey}. Dedicated workshops on computational approaches to historical language change took place at the ACL conferences \cite{change_workshop:2019,lchange-2021-international,lchange-2022-approaches} and the results of the SemEval-2020 Task 1 on unsupervised lexical semantic change detection were announced in March 2020 \cite{schlechtweg2020semeval}. Shared tasks for other languages followed soon \cite{basile2020diacr,rushifteval2021}. 

The majority of the SemEval-2020 shared task participants employed methods based on word embeddings of various types. About half of them tried to make use of contextualized (`token-based') architectures like ELMo \cite{elmo2018} or BERT \cite{devlin2019bert}. Although the winning systems still used non-contextualized (`static' or `type-based') embeddings like word2vec \cite{Mikolov_representation:2013}, the difference in scores was not dramatic and we are most likely going to see more work in this direction. We agree with \newcite{schlechtweg2020semeval} that as the contextualizing technologies mature, there will be a better understanding of how to properly use them for semantic change related tasks. Indeed, at the RuShiftEval shared task on LSCD for Russian \cite{rushifteval2021}, the leader-board was already dominated by contextualized models.

The current paper aims to contribute to this improved understanding by qualitatively analyzing the output of contextualized embedding-based approaches to the diachronic semantic change detection task. Hence, our work falls into the second category of ground truth semantic change evaluation, as defined by \newcite{hengchen2021challenges}: what is evaluated is the ranked output of the methods under investigation.

We here focus on Subtask~2 of SemEval-2020 Task~1: to rank a list of words by the degree of their semantic change between two historical corpora belonging to different time bins. The submissions were evaluated by their Spearman rank correlation against human annotations. This task was offered for four languages, each with their own word list and corpora: English, German, Latin and Swedish. One of the submissions in this Subtask was delivered by the UiO-UvA team \cite{kutuzov2020uio}. It used pre-trained ELMo models and achieved the average score of 0.37 at the evaluation phase (the second best contextualized embedding-based system in this phase), and 0.62 at the post-evaluation phase (the best result overall in this phase). We chose their methods for closer inspection, because the implementations were publicly available, and the methods themselves are quite typical for the semantic change detection field (see below).

The contributions of this paper are twofold:
\begin{enumerate}
\item We propose a simple improvement to the approach in \newcite{kutuzov2020uio} by ensembling two of their best-performing methods. We show that it avoids the necessity to decide what method to choose, while still outperforming strong baselines.
\item We qualitatively examine the output of the contextualized methods for semantic change detection in English. We analyze examples of both correct and incorrect cases of detected semantic change.  The latter findings are arguably more important for future studies, as one learns on errors. We propose a categorization of such problematic cases, relating them to inherent properties of pre-trained contextualized architectures in particular and distributional approaches in general.
\end{enumerate}

\section{Contextualized methods for detecting semantic change}

Two methods for estimating semantic change were proposed in \newcite{kutuzov2020uio}: PRT and APD (further detailed below). The methods are architecture-agnostic and can be used with any model able to produce contextualized token representations for a given sequence of word tokens. Overall, these methods can be considered typical representatives of using contextualized word embeddings for the task of semantic change detection: they boil down to directly comparing token embeddings of the target word in two periods; see \cite{martinc-etal-2020-leveraging} for a similar technique. Another possible approach (which we hope to analyze in the future) is clustering token embeddings into groups loosely corresponding to word senses and then comparing their time-specific distributions \cite{martinc2020capturing,cuba-gyllensten-etal-2020-sensecluster,giulianelli-etal-2020-analysing}.

The common part of both the PRT and APD methods is as follows. Given two time periods $t_1$ and $t_2$, two corresponding corpora $C_1$ and  $C_2$, and a set of target words, a language model (regardless of what it has been pre-trained on) is used to obtain contextualized token embeddings\footnote{Representations from the top layer of the model were used, since they yielded the best results according to \newcite{kutuzov2020uio}.} of each occurrence of the target words in $C_1$ and $C_2$. Each target word $w$ is then represented by two `usage matrices' $\textbf{U}_w^{t_1}$ and $\textbf{U}_w^{t_2}$ consisting of all token embeddings produced for $w$. A \textit{change score} is computed from these matrices, indicating the degree of semantic change undergone by a word between $t_1$ and $t_2$. The target words are ranked by this  value. The methods differ in how exactly change scores are computed:

\begin{itemize}
    \item \textbf{Inverted cosine similarity over word prototypes (PRT)}: the degree of change for $w$ is calculated as the inverted cosine similarity between the average token embeddings (`prototypes') of all $w$  occurrences in $\textbf{U}_w^{t_1}$ and $\textbf{U}_w^{t_2}$ correspondingly:
\begin{align}
    \operatorname{PRT}\left(\textbf{U}_w^{t_1}, \textbf{U}_w^{t_2}\right) = \frac{1}{c\left(\frac{\sum_{\textbf{x}_i \in \textbf{U}_w^{t_1}} \textbf{x}_i}{N_w^{t_1}}, \frac{\sum_{\textbf{x}_j \in \textbf{U}_w^{t_2}} \textbf{x}_j}{N_w^{t_2}}\right)}  
\end{align}
where $N_w^{t_1}$ and $N_w^{t_2}$ are the numbers of occurrences of $w$ in time periods $t_1$ and $t_2$, and $c$ is a similarity metric, for which we use cosine similarity. High PRT values indicate a higher degree of semantic change. 

\item \textbf{Average pairwise cosine distance between token embeddings (APD)}: the degree of  change for $w$ is measured as the average distance between all possible pairs of token embeddings in $\textbf{U}_w^{t_1}$ and $\textbf{U}_w^{t_2}$:
\begin{align}
    \operatorname{APD}\left(\textbf{U}_w^{t_1}, \textbf{U}_w^{t_2}\right) =  \frac{1}{N_w^{t_1} \cdot N_w^{t_2}} \sum_{\textbf{x}_i \in \textbf{U}_w^{t_1},\ \textbf{x}_j \in \textbf{U}_w^{t_2}} d\left(\textbf{x}_i, \textbf{x}_j\right)
\end{align}
where $d$ is the cosine distance ($1 - c$ where $c$ is cosine similarity). High APD values indicate a higher degree of semantic change. 
\end{itemize}

\newcite{kutuzov2020uio} report that different test sets from the shared task manifested strong preference for either the PRT or the APD method, and that this is correlated with the distribution of gold scores in the test set (but not with its language). If the right method was chosen, then using contextualized embeddings to rank words by their degree of semantic change consistently outperformed the shared task baselines (frequency-based and count-based approaches) and the methods relying on type-based embeddings with orthogonal alignment \cite{hamilton2016cultural}.

\begin{table*}
    \centering
    \begin{tabular}{l|lllll|l}
\textit{Method} & \textit{English} & \textit{German} & \textit{Latin} & \textit{Swedish} & \textit{GEMS} & \textit{Average} \\
\midrule
\multicolumn{7}{c}{\textbf{SemEval-2020 Task 1 baselines}}  \\
\midrule
FD (frequency difference) & -0.217 & 0.014 & 0.020 & -0.150 & 0.068 & 0.094  \\
CNT+CI+CD (count-based) & 0.022 & 0.216 & 0.359* & -0.022 & 0.256* & 0.166  \\   
\midrule
\multicolumn{7}{c}{\textbf{Cosine distance with static  embeddings (word2vec)}}  \\
Orthogonal Procrustes alignment & 0.285 & 0.439* & 0.387* & 0.458* & 0.235* & 0.361  \\
\midrule
\multicolumn{7}{c}{\textbf{Contextualized embeddings}}  \\
\midrule
\textit{BERT} PRT & 0.225 & 0.590* & \textbf{0.561}* & 0.185 & \textbf{0.394}* & 0.391  \\
\textit{BERT} APD & 0.546* & 0.427* & 0.372* & 0.254 & 0.243* & 0.368  \\
\textit{BERT} PRT/APD & 0.498* & 0.537* & 0.431* & 0.267 & 0.332* & 0.413 \\
\midrule
\textit{ELMo} PRT & 0.254 & \textbf{0.740}* & 0.360* & 0.252 & 0.323* & 0.386  \\
\textit{ELMo} APD & \textbf{0.605}* & 0.560* & -0.113 & \textbf{0.569}* & 0.323* & 0.389  \\
\textit{ELMo} PRT/APD & 0.546* & 0.678* & 0.036 & 0.546* & 0.360* & \textbf{0.433} \\
\midrule
\midrule
\multicolumn{7}{c}{\textbf{Inter-correlations between \textit{ELMo} PRT and APD predictions }}  \\
Spearman's $\rho$  & 0.589*  & 0.655*  & 0.423*  & 0.538*  & 0.319*  & 0.505  \\
Pearson's r & 0.547*  & 0.656*  & 0.589*  & 0.698*  & 0.495*  & 0.597 \\
\bottomrule
    \end{tabular}
    \caption{Spearman correlation with the gold standard per test set for the best methods from \cite{kutuzov2020uio} and our PRT/APD ensemble approach. `*' denotes statistical significance of the correlation as measured by the two-sided p-value, $p < 0.05$.} 
    \label{tab:test_task2_best}
\end{table*}

However, in a realistic setting it is obviously problematic to assume knowledge of the statistical properties of the target words beforehand. So, how should one choose between the PRT and APD methods? We found that simply averaging the PRT and APD estimates yields very robust predictions. In Table~\ref{tab:test_task2_best}, we reproduce the results from \newcite{kutuzov2020uio}, including the word2vec baseline, and add the `PRT/APD' row with the scores we got using the ensemble approach.  Note that in addition to the 4 shared task test sets, we also report results on the GEMS semantic change test set for English \cite{baroni:2011}. For individual test sets, the performance of PRT/APD usually lies in between PRT and APD, but when averaged over all five test sets, it ranks higher than any individual method, and this effect holds for both ELMo and BERT, with the best result yielded by ELMo. When compared to the shared task leader-board \cite{schlechtweg2020semeval}, the PRT/APD + ELMo combination outperforms all contextualized embedding-based systems in Subtask 2, supporting the same observation in \cite{kutuzov2020uio}. 

Thus, the APD and PRT methods are complimentary, although their predictions are strongly correlated (see the bottom of Table~\ref{tab:test_task2_best}). Together they act as a top-performing ensemble of the models, with the additional benefit of not having to worry about what method to choose. In the rest of this paper, we will use the PRT/APD method to produce semantic change scores for qualitative analysis. Note that since these scores are produced by an ensemble model, they are less interpretable than the original separate PRT and APD values. However, a manual inspection showed that the separate methods yield the same categories of errors as the combined score; see Section~\ref{sec:shift_bad} below.

\section{Data and models used}
For our in-depth analysis of the results, we use textual data from the Corpus of Historical American English or COHA \cite{coha}
(it is certainly desirable to reproduce this analysis for other languages, which we leave for future work). 
In particular, we deal with 5 COHA sub-corpora corresponding to five decades: the 1960s, 1970s, 1980s, 1990s and 2000s. Note that this setup is slightly different from the SemEval-2020 Task 1 in that we have a sequence of five time bins. With this, we aim to trace the lasting evolution of word meaning, not limited to changes between two time periods. The employed time span means we deal with relatively short-term meaning changes.

We chose ELMo as a contextualizer based on its better performance (Table~\ref{tab:test_task2_best}) and much lower computational requirements than BERT. It allowed us to train a single model from scratch on the concatenation of all COHA texts belonging to the five decades mentioned above (the full corpus size is about 127 million word tokens, and we trained for 5 epochs).
The texts were tokenized and lemmatized with the English UDPipe tagger trained on the Universal Dependencies 2.3 treebank \cite{udpipe:2017}, discarding punctuation marks and lower-casing all words.

The list of words to analyze is a concatenation of all words from the SemEval-2020 Task 1 English test set, all words from the GEMS test set, and 1000 randomly sampled words occurring in all five COHA sub-corpora with frequency in each sub-corpus higher than 100. After excluding numerals, function words and the words with a total frequency of  less than 1000 occurrences across all decades (to discard unstable representations of rare words), the resulting word list contains 690 entries. For each of them, we used our ELMo model to calculate their PRT/APD scores in the four consecutive pairs of the COHA decades (1960s--1970s, 1970s--1980s, 1980s--1990s, 1990s--2000s), thus producing a score matrix $\mathbf{M} \in \mathbb{R}^{690\times4}$. Below we examine the actual scores in this matrix, and how they are related to processes in the recent history of English.

\section{Well-behaved examples} \label{sec:shift_good}

For many words, the scores do signal real changes, like a new emergent sense. Let us consider the word \word{cell} as an example.  The dataset from \newcite{tsakalidis2019mining}, based on the Oxford English Dictionary definitions, mentions it as having acquired a new sense of \sense{mobile phone} after 2000. Recall that PRT/APD produces as an output a measure of how strong the semantic change of a target word was between two time bins; this measure characterizes a pair of decades in our case. \word{Cell} received a change coefficient of 0.673 for the 1960--1970 pair (arguably corresponding to the start of its widespread usage in the biological sense). 

After that, the estimated degrees of change were smaller, with 0.669 for 1970--1980s and 0.672 for 1980--1990s. However, 1990--2000s had a change coefficient of 0.695 (the highest for this word across all decades), most likely reflecting the new \sense{mobile phone} sense. As a side note, it might look like the PRT/APD values show very little variation: in fact the average standard deviation of $\mathbf{M}$ values across four time period pairs is $0.04$, with the average PRT/APD value being about $0.70$. This means that the change coefficients for \word{cell} are actually lower than the mean value in our dataset (z-score of 0.695 is $-0.17$). See more on this in the next Section~\ref{sec:shift_bad}.

\begin{figure}
    \centering
    \includegraphics[width=0.45\linewidth]{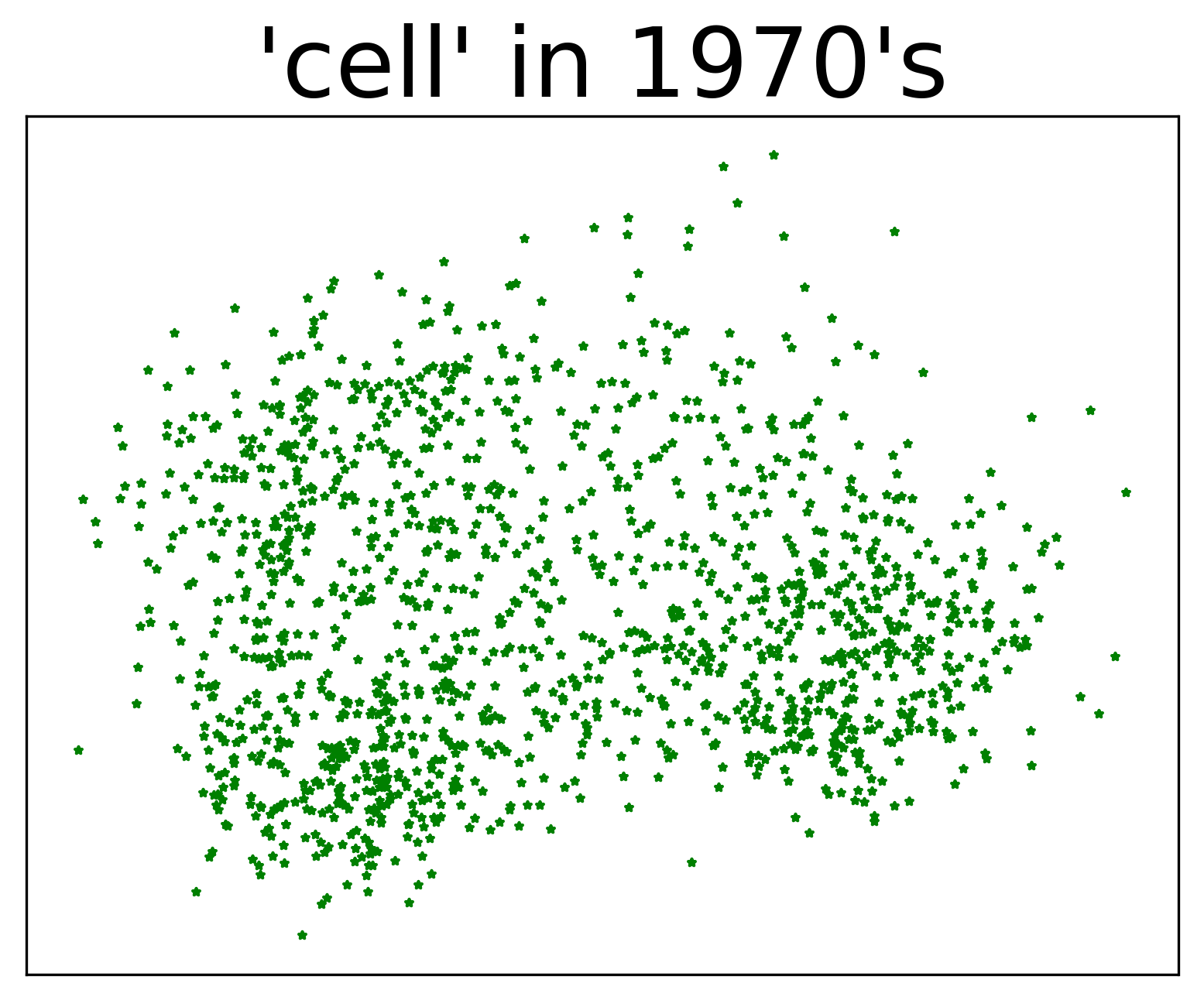}
    \includegraphics[width=0.45\linewidth]{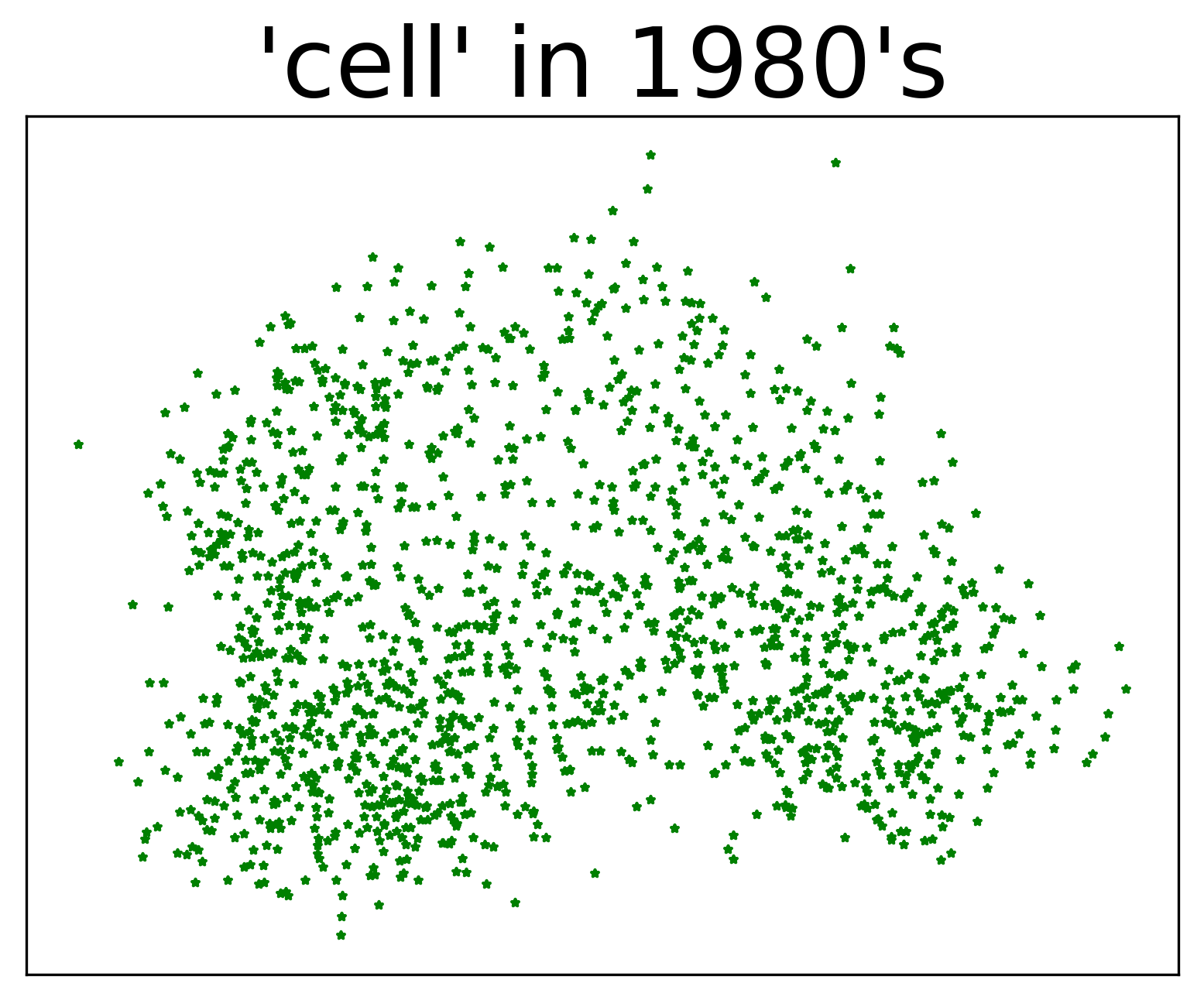}
    \includegraphics[width=0.45\linewidth]{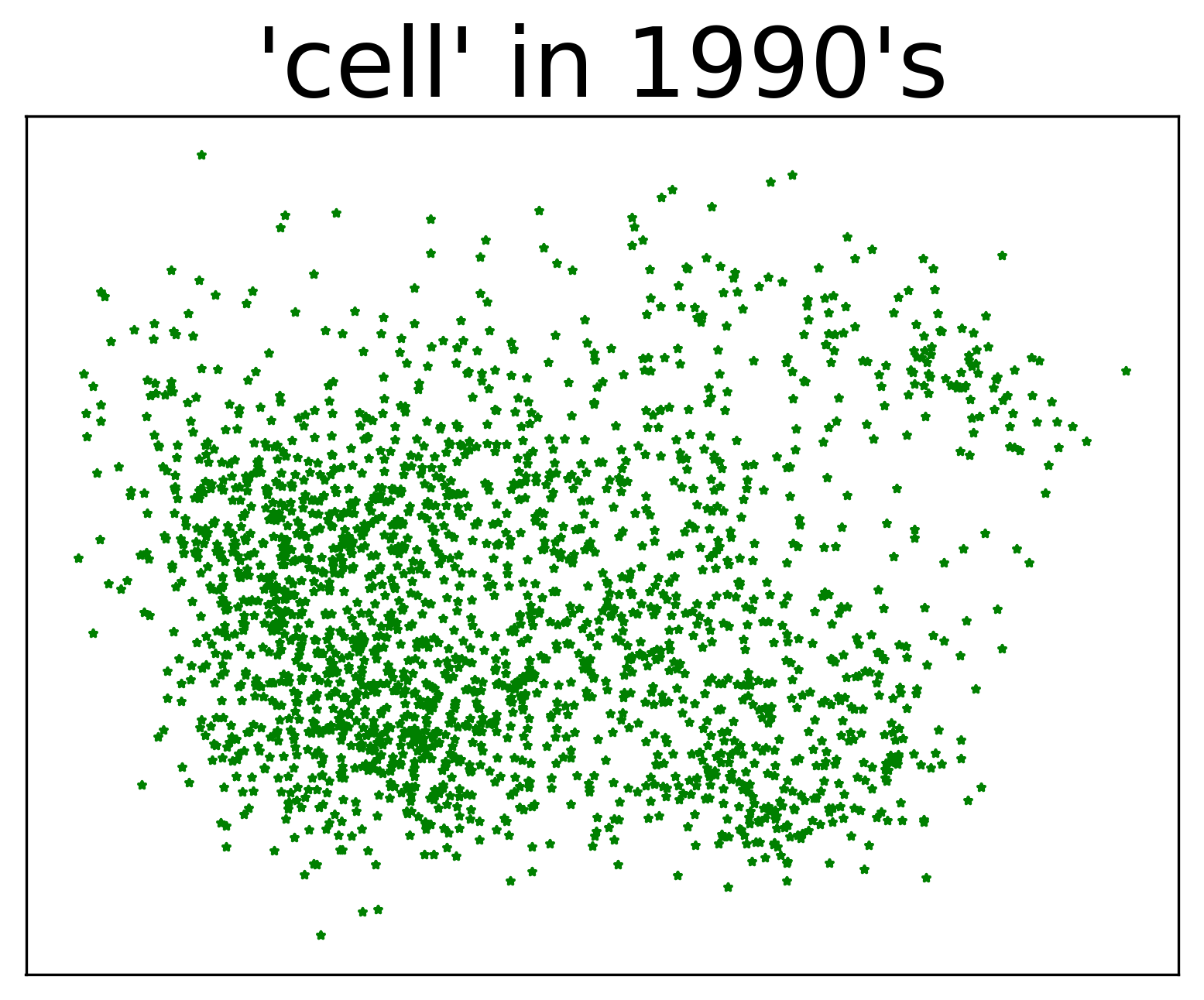}
    \includegraphics[width=0.45\linewidth]{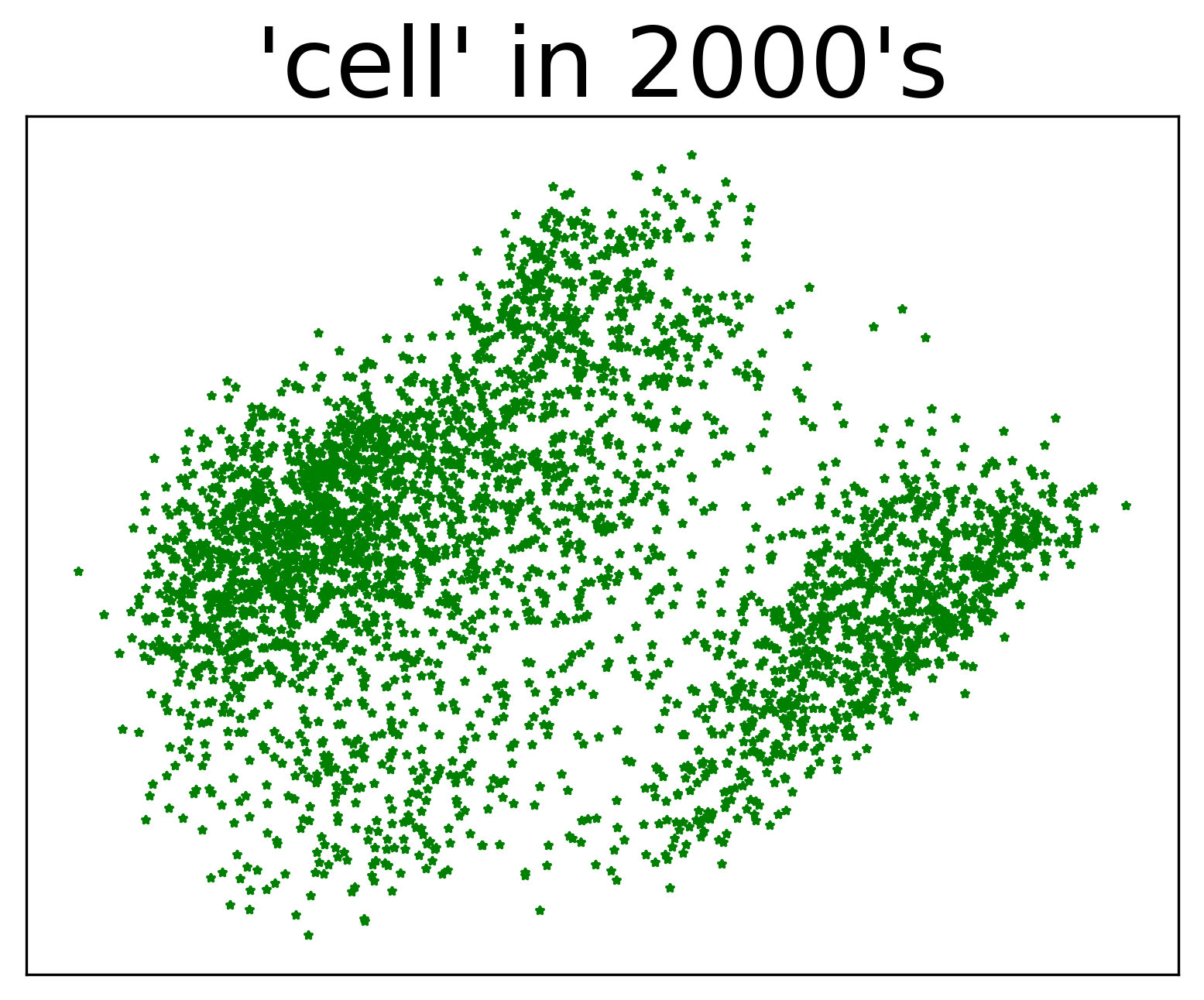}
    \caption{PCA projections of contextualized token embeddings of \word{cell} in four different decades.}
    \label{fig:cell_dia}
\end{figure}

Unlike the static word embedding approaches, using contextualized models allows one to visually explore the individual occurrences of a given word in  different senses. For this purpose, we use Principal Component Analysis (PCA) to reduce the contextualized token embeddings of \word{cell} in our diachronic sub-corpora to their 2-dimensional projections. Figure \ref{fig:cell_dia} shows these projections for the decades from the 1970s through the 2000s.

Even at a glance, it is possible to see that in the 2000s, some radical changes in the groupings of the \word{cell} token embeddings occurred. The three previous decades are all characterized by a rather vague separation of this word's usages into two clusters (at the left and at the right part of the vector space). In the 2000s, we observe the appearance of a new cluster: now there are two strong clusters to the left and a third one to the right. But what senses do these clusters correspond to? Fortunately, since each point on the plot represents a particular \word{cell} occurrence from a particular decade's sub-corpus, we can retrieve their corpus contexts and manually inspect them. Of course, we did not inspect \textit{all} occurrences: both due to their amount (thousands) and due to the absence of clear-cut cluster boundaries. Instead, we randomly sampled about 20 occurrences from the core area of each apparent cluster and examined them.\footnote{The same method is used below throughout the paper. In all cases when visible clusters appeared in the projections, they were strongly consistent, with at least 90\% of the randomly sampled data points in a cluster belonging to a particular sense.}

We observe that in the 1970s, 1980s and 1990s, the right-hand cluster mostly contains sentences with \word{cell} in the sense of \sense{prison cell}, see example \ref{ex:prison}:
\begin{covexample}
\begin{enumerate}
    \item `I'd known Archie Meltzer, the chief turnkey on duty, for over ten years, but you wouldn't have known it from the way he processed me for the \textbf{cells}.'
    \item `It also happened to me in a jail \textbf{cell}.'
    \item `If she had been writing to somebody in the darkness of her prison \textbf{cell}, what had she done with the message?'
\end{enumerate}
\label{ex:prison}
\end{covexample}

The left cluster (stably increasing its relative size over time) mostly contains sentences with \word{cell} in the biological sense, with examples given in \ref{ex:bio}.

\begin{covexample}
\begin{enumerate}
    \item `The sexual \textbf{cells} of Pyronema show this in ascomycetes.'
    \item `It's how a \textbf{cell} decides whether it becomes a muscle \textbf{cell} or a skin \textbf{cell}.'
    \item `If those \textbf{cells} are found to be cancerous after being sent to a lab, that's a definite diagnosis.'
\end{enumerate}
\label{ex:bio}
\end{covexample}

\begin{figure}
    \centering
    \includegraphics[width=\linewidth]{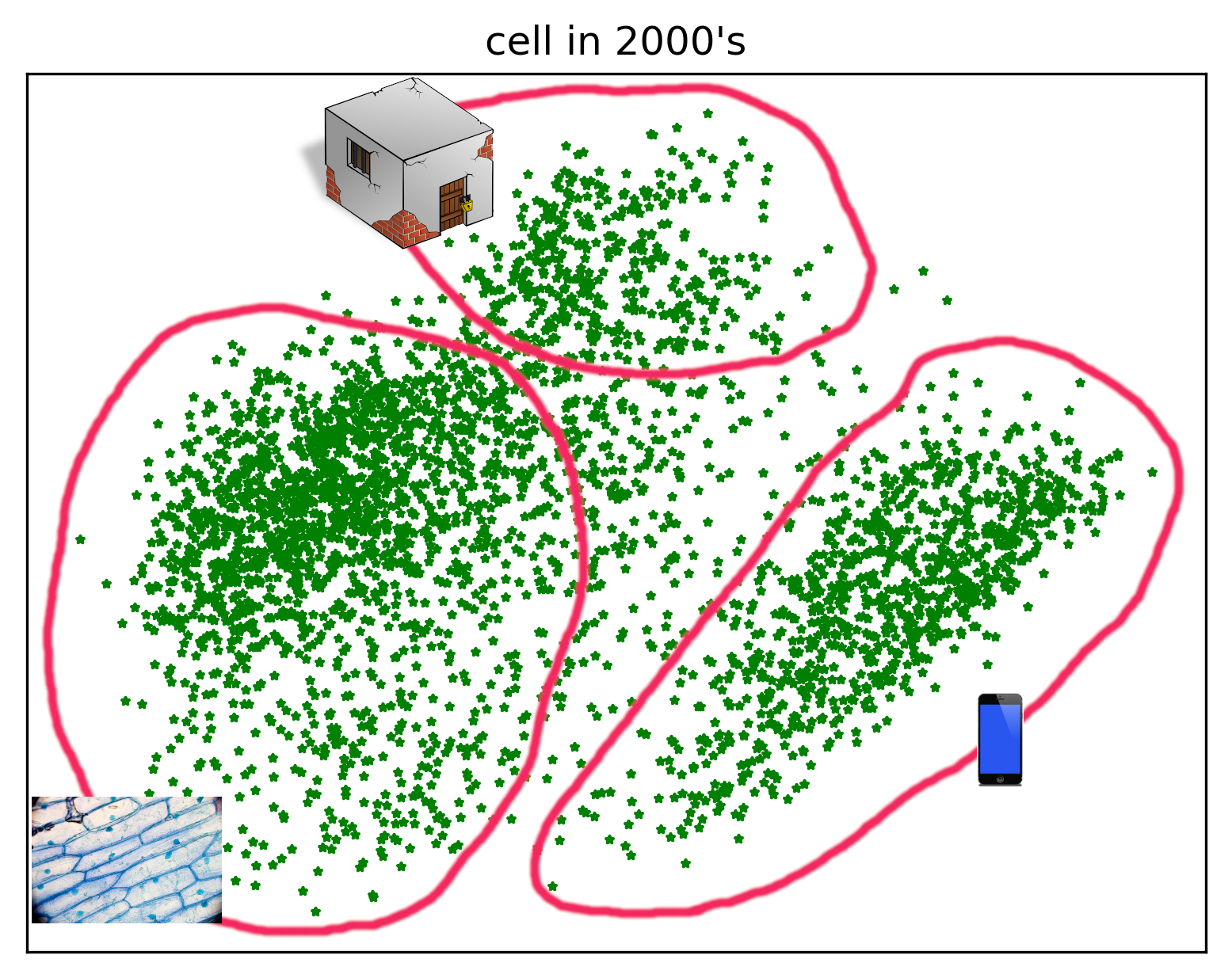}
    \caption{PCA projection of ELMo token representations of each occurrence of the word \word{cell} in the 2000s, with clusters labeled with senses.}
    \label{fig:elmo_cell_labeled}
\end{figure}

After exploring the points in the 2000s plot in the same way, one observes that the two clusters on the left correspond to the old senses of \word{cell} (biological still at the bottom and prison at the top).  
But the new large cluster on the right almost exclusively consists of sentences mentioning \word{cell} in the sense of \sense{mobile phone} (see examples in \ref{ex:phone} and Figure \ref{fig:elmo_cell_labeled} displaying these clusters with labels).

\begin{covexample}
\begin{enumerate}
   \item `But how well do the service providers fulfill that objective, and what about the other health and safety risks - exposure to radio waves and potentially fatal driver distraction - that the growing use of \textbf{cell} phones raise?'
    \item `...he walked past, nearly dislodging the \textbf{cell} phone she had balanced between her chin and her left shoulder.'
    \item `You still have the same \textbf{cell} number.'
\end{enumerate}
\label{ex:phone}
\end{covexample}

\begin{figure}
    \centering
    \includegraphics[width=\linewidth]{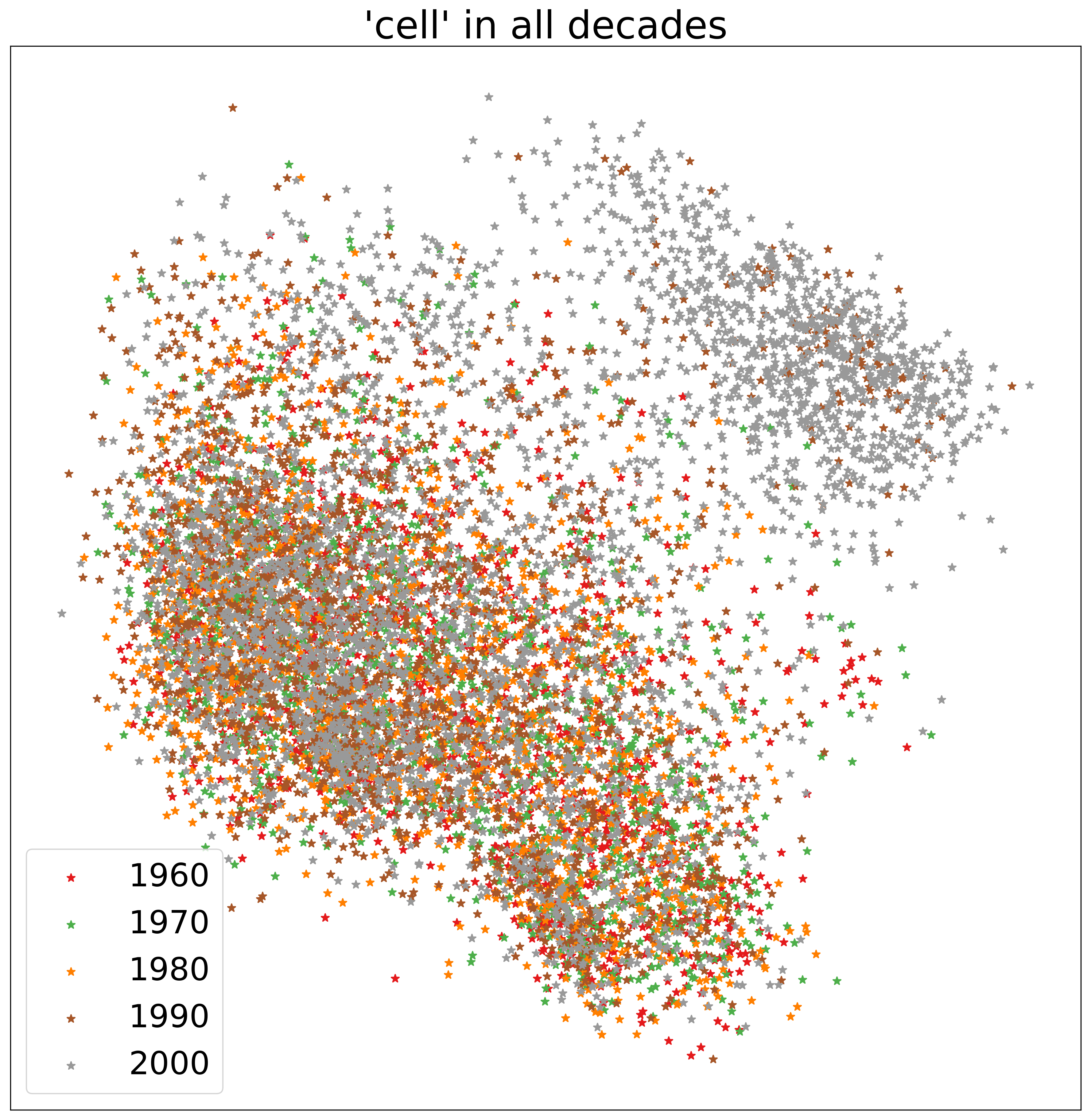}
    \caption{PCA projection of contextualized token embeddings of \word{cell} in all 5 COHA decades. Colors correspond to time periods.}
    \label{fig:elmo_cell_all}\vspace{-1ex}
\end{figure}

One can also visualize token embeddings for \word{cell} across all five time bins in one plot, as shown in Figure~\ref{fig:elmo_cell_all}. Here, PCA dimensionality reduction is performed for all occurrences of this word (about 7500 total), and thus we can see how usages from different decades (shown in different colors) are grouped in relation to each other. The top right cluster is inhabited almost exclusively with the occurrences from the 2000s and to a less extent the 1990s. Not surprisingly, it contains sentences where \word{cell} is used in the \sense{mobile phone} sense. At the same time, in other parts of the plot, occurrences from all decades are distributed more or less uniformly, supporting our previous observation that in the 1960s, 1970s and 1980s, this word did not experience significant semantic changes.

In the case of \word{cell}, the groupings of contextualized representations and the detected changes are undoubtedly connected to a new sense emerging (thus, a diachronic semantic shift). The relations between different senses of \word{cell} fall into the category of \textit{homonymy}, where word senses are not directly related to each other (at least, synchronically). However, one can trace the cases of \textit{polysemy} as well, where senses are synchronically related to each other. As an example, let us look at the adjective \word{virtual}. It experienced its strongest change of $0.769$ in the 1980--1990 pair (its z-score is $1.9$ in the full $\mathbf{M}$).

Before 1990s, \word{virtual} was used mostly in two closely related senses: \sense{being such in essence or effect though not formally recognized or admitted} (major one) and \sense{related to a hypothetical particle whose existence is inferred from indirect evidence} (minor).\footnote{The definitions are taken from the Merriam-Webster dictionary (\url{https://www.merriam-webster.com/}).} However, the 1990s saw the emergence of a large number of \word{virtual} usages in the sense of \sense{simulated on a computer or computer network}, especially in the expression `virtual reality' (almost one third of all usages). This sense is related to the previous ones, thus manifesting a case of polysemy. 
The emergence of a new related sense in the 1990s is captured by contextualized embedding based methods, producing a higher change score for this time bin in comparison to the previous 1980s decade. 
We can also observe a much weaker change score of 0.740 in the 1990--2000 pair. The manual inspection of the occurrences shows that in the 2000s, \word{virtual} was still used a lot in this new third sense (interestingly, the `virtual reality' expression itself almost came out of usage, constituting now only 6\% of all \word{virtual} occurrences). 

\begin{figure}
    \centering
    \includegraphics[width=\linewidth]{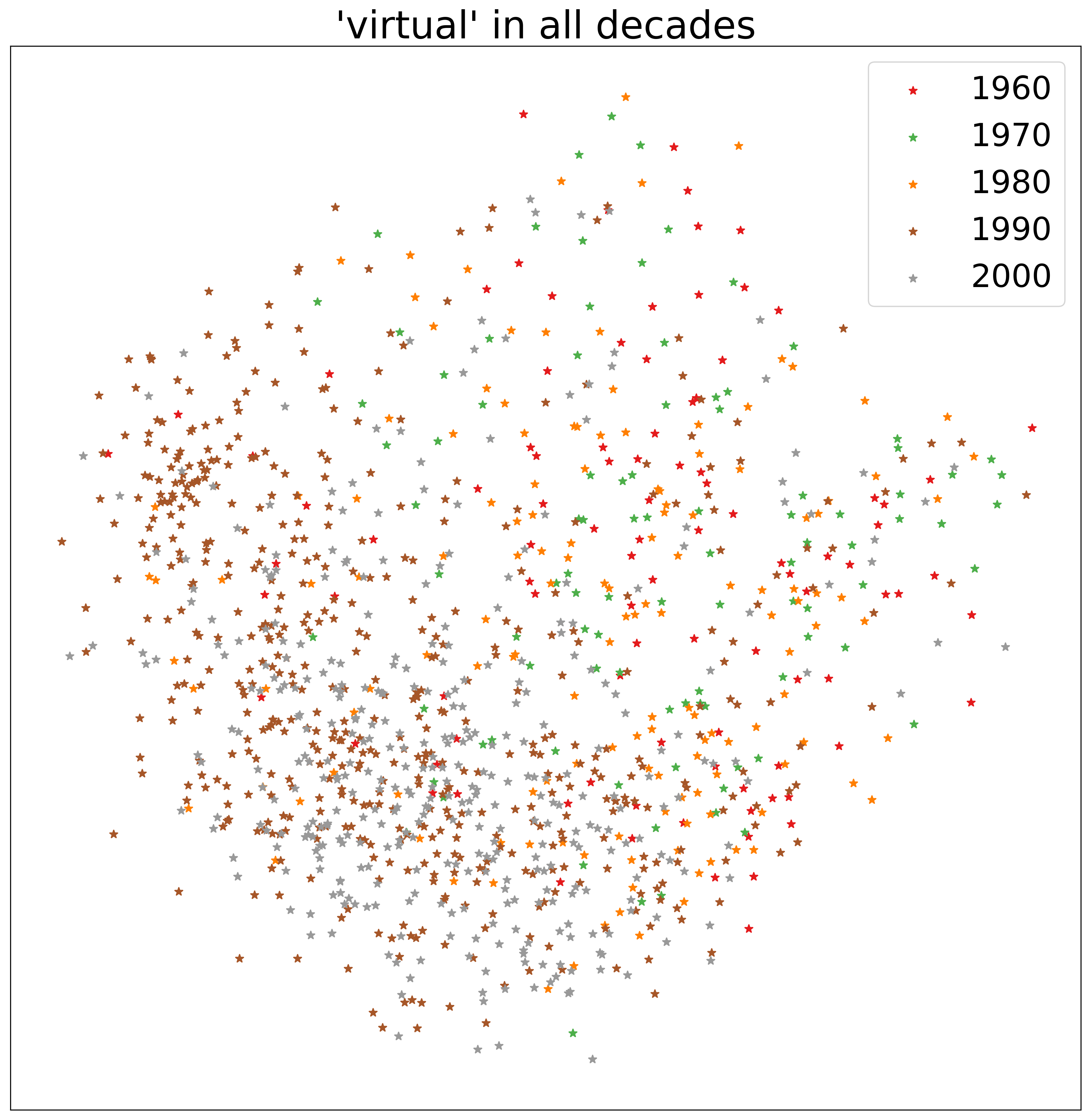}
    \caption{PCA projection of contextualized token embeddings of \word{virtual} in all 5 COHA decades.}
    \label{fig:elmo_virtual_all}
\end{figure}

On the plot of \word{virtual} token embeddings across five COHA decades (Figure~\ref{fig:elmo_virtual_all}), the \sense{simulated on a computer or computer network} usages occupy the left part of the plot, with the `virtual reality' phrases concentrated in the left top corner (as confirmed by manual inspection). The left part contains almost exclusively the occurrences from the 1990s and from the 2000s, while the left top corner is dominated by the 1990s.

So far so good: the contextualized embedding-based methods not only demonstrate high performance on the evaluation sets, they also produce interpretable predictions corresponding to well-known diachronic semantic shifts. But let us also look at a darker side of the $\mathbf{M}$ score matrix.

´´´


\section{Problematic  examples} \label{sec:shift_bad}

The picture is not as clear if one gets beyond hand-picked well-behaved examples. As mentioned above, the change coefficient of \word{virtual} when comparing the 1990s to the 1980s was 0.769. But the absolute values (and even z-scores) here are not very informative. There is no well-defined threshold: it is not the case that the change coefficients higher than, say, $0.7$ always correspond to some breaking points in the word evolution. There are much stronger bursts which do not yield to such an explanation. Table \ref{tab:points_change} lists 10 words with the highest change coefficients in $\mathbf{M}$. As can be seen, these changes are indeed unusually strong, all of them being more than $2$ standard deviations away from the mean change score. However, none of them can be immediately interpreted as acquiring or losing a sense. What is the cause of these bursts?

\begin{table}
    \centering
    \begin{tabular}{l|c|c|c}
    \textbf{Word} & \textbf{Decade pair} & \textbf{Change} & \textbf{z-score}\\
    \midrule
    \textcolor{blue}{\word{banish}} & 1980s--1990s & 0.794 & 2.60 \\  
    \textcolor{red}{\word{designate}} & 1980s--1990s & 0.792 & 2.54 \\  
    \textcolor{magenta}{\word{mg}} (m/gram) & 1980s--1990s & 0.791 & 2.52 \\   
    \textcolor{red}{\word{progressive}} & 1990s--2000s & 0.782 & 2.27  \\  
    \textcolor{magenta}{\word{indirectly}} & 1990s--2000s & 0.780 & 2.21 \\  
    \textcolor{red}{\word{form}} & 1990s--2000s & 0.780 & 2.21 \\  
    \textcolor{red}{\word{subsequently}} & 1980s--1990s & 0.780 & 2.21  \\  
    \textcolor{magenta}{\word{neutral}} & 1990s--2000s & 0.779 & 2.18 \\  
    \textcolor{brown}{\word{traditionally}} & 1990s--2000s & 0.779 & 2.18 \\ 
    \textcolor{red}{\word{pointed}} & 1960s--1970s & 0.778 & 2.16 \\  
    \bottomrule
    \end{tabular}
    \caption{10 points with the highest change scores in 5 decades of COHA (as measured by PRT/APD). Z-scores are computed on the full $\mathbf{M}$. Word color indicates its class, see Section~\ref{sec:shift_bad}.}
    \label{tab:points_change}
\end{table}

\subsection{Categories of problematic examples}
Indeed, none of the 10 words with the highest scores is a schoolbook example of a semantic shift. We emphasize it does not necessarily imply outright errors or `false positives'. As we show below, a good part of these words in fact do have reasons to be assigned high change scores; it is just that these reasons are somewhat different from what a historical linguist would expect to see. 

Looking closely at these cases reveals three general word classes which trigger high semantic change score as measured by the PRT/APD approach, but at the same time did not undergo any semantic shifts in the classic understanding of the term \cite{bloomfield}. The classes are (colors correspond to those in Table~\ref{tab:points_change}):

\begin{enumerate}
    \item \textcolor{red}{Words of strongly context-dependent meaning} (\word{designate}, \word{progressive}, etc): their token embeddings are very different from each other (and thus change scores are high) when compared either synchronically or diachronically.
    \item \textcolor{magenta}{Words frequently used in a very specific context in a particular time bin}, different from other periods (\word{mg}, \word{indirectly}, etc). It can be looked at either as a result of (unintended) domain shifting when building a corpus or as contextual variance which really exists in language, but did not yet lead to the emergence of a new lexicographic sense (or losing an old one). Note that \newcite{shoemark-etal-2019-room} observed very similar phenomena when analyzing Twitter data with static word embeddings. We will also call such cases `data bursts'. There is an interesting sub-type of this class:
    \begin{itemize}
        \item \textcolor{blue}{words used as a proper name in a particular time bin} (\word{banish}, etc.); this leads to extremely high contextual variance and the emergence of isolated token clusters.
    \end{itemize}
    \item \textcolor{brown}{Words undergoing syntactic changes, not semantic ones}; see below.
\end{enumerate}

Note that the assignment of data points to classes in Table~\ref{tab:points_change} was not done as a part of a full-fledged annotation effort with pre-defined error categories. Rather, this is a product of qualitative error analysis conducted by the authors: that is, the classes were identified as an attempt to group and systematize the problematic predictions of the methods used. We by any means do not claim that this grouping is the only one possible; however, as shown below, it models the data well enough to produce meaningful insights.

We remind the reader that the change coefficients were produced by the ensemble PRT/APD method. However, the PRT and APD methods on their own suffer from the same categories of problems. We analyzed 10 words with the highest estimated degree of change for the separate methods as well, and found them to largely overlap with those produced by  PRT/APD; see Table~\ref{tab:points_change_separate}. For APD, 60\% of the points are the same words as for PRT/APD, for PRT it is 20\%, but these two words are at the top of the list.\footnote{Spearman $\rho$ correlation between predictions of APD and PRT on $\mathbf{M}$ varies from $0.19$ to $0.34$, depending on a particular pair of decades; for Pearson, it is from $0.13$ to $0.16$; all the correlations are statistically significant.}

\begin{table}
    \centering
    \begin{tabular}{l|l|l|l}
    \textbf{PRT} (score)  & \textbf{Bin} & \textbf{APD} (score) & \textbf{Bin} \\
    \midrule
    \textcolor{magenta}{\word{mg}} (1.17) & 1990s & \textcolor{red}{\word{designate}} (0.57) & 2000s \\  
    \textcolor{blue}{\word{banish}} (1.11) & 1990s & \textcolor{red}{\word{progressive}} (0.56) & 2000s \\ 
    \word{don} (1.11) & 1980s & \textcolor{red}{\word{form}} (0.56)  & 2000s \\   
    \textcolor{blue}{\word{crunch}} (1.07) & 1970s & \textcolor{red}{\word{subsequently}} (0.55) & 1970s  \\ 
    \word{immune} (1.07) & 1980s & \textcolor{red}{\word{lead}} (0.55) & 1990s \\  
    \textcolor{red}{\word{clayton}} (1.07) & 1970s & \textcolor{brown}{\word{traditionally}} (0.55)  & 2000s  \\  
    \textcolor{blue}{\word{norm}} (1.06) & 1970s & \textcolor{red}{\word{pointed}} (0.55) & 1970s  \\ 
    \textcolor{magenta}{\word{brian}} (1.06) & 1970s & \textcolor{red}{\word{truly}} (0.55) & 2000s  \\  
    \textcolor{magenta}{\word{ian}} (1.06) & 1980s & \textcolor{red}{\word{mere}} (0.55) & 2000s  \\ 
    \textcolor{magenta}{\word{sequence}} (1.06) & 2000s & \textcolor{red}{\word{savage}} (0.55) & 2000s \\  
    \bottomrule
    \end{tabular}
    \caption{10 points of the strongest change in 5 decades of COHA, as measured separately by PRT and APD. Word color indicates its class, see Section~\ref{sec:shift_bad}. `Bin' columns denote the decade when the change occurred.}
    \label{tab:points_change_separate}
\end{table}

An interesting observation is that each separate method tends to `favor' different classes of problematic examples: while for PRT, seven words out of the top 10 are cases of \textcolor{magenta}{data bursts} (including the \textcolor{blue}{proper name} subclass), for APD, nine of the top 10 are  \textcolor{red}{words with strongly context-dependent meaning}. The PRT/APD method yields a more balanced distribution of these two classes (each takes approximately half of the top 10 list): this is arguably one of the reasons for its higher empirical performance. This aligns well with the assumption about the complementary nature of PRT and APD that we already mentioned before. The analysis of the reasons for this behavior is an interesting topic for future studies.

As a side note, two words predicted as changed by the PRT method do not fall into any of our categories: \word{don} and \word{immune}. \word{Don} stems from what seems to be a corpus pre-processing issue on the COHA side: in the 1980s sub-corpus of COHA, the frequency of \word{don't} tokenized as \word{don ' t} (with two spaces) is two orders of magnitude higher than in the other decades. This leads to the appearance of a very distinct \word{don} cluster in this time bin. For \word{immune}, we observe that in the 1980s, it starts being actively used in the phrase \word{immune system}, again forming a separate cluster. This is not a temporary data burst, since it continued in the 1990s and in the 2000s. The dynamics of \word{immune} is arguably related to the discovery of the HIV virus in the beginning of the 1980s, and thus, it can (cautiously) be acknowledged as a well-behaved example, not a problematic one. But let us return to the PRT/APD predictions.  

\begin{figure}
    \centering
    \includegraphics[width=0.48\linewidth]{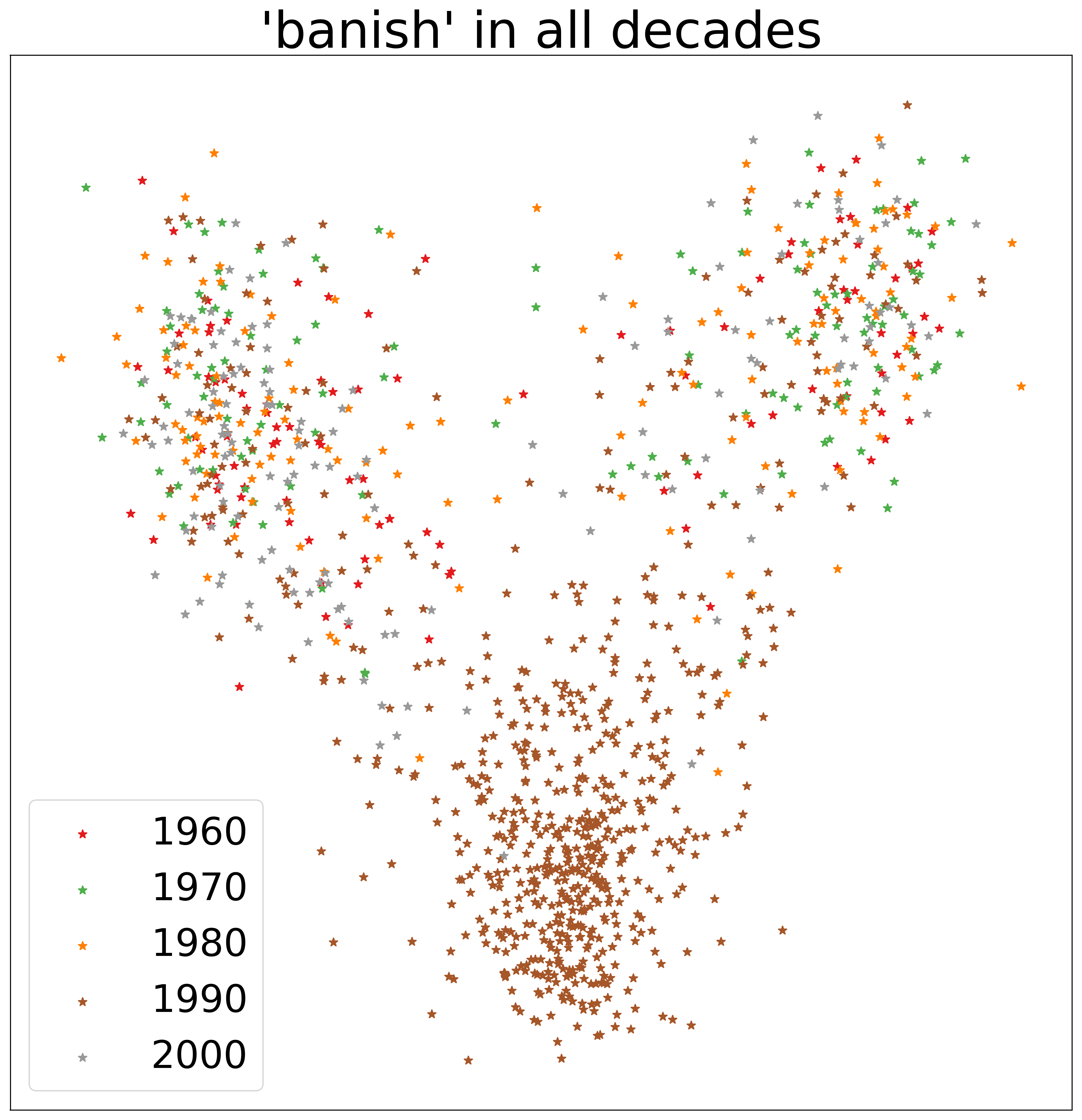}
    \includegraphics[width=0.48\linewidth]{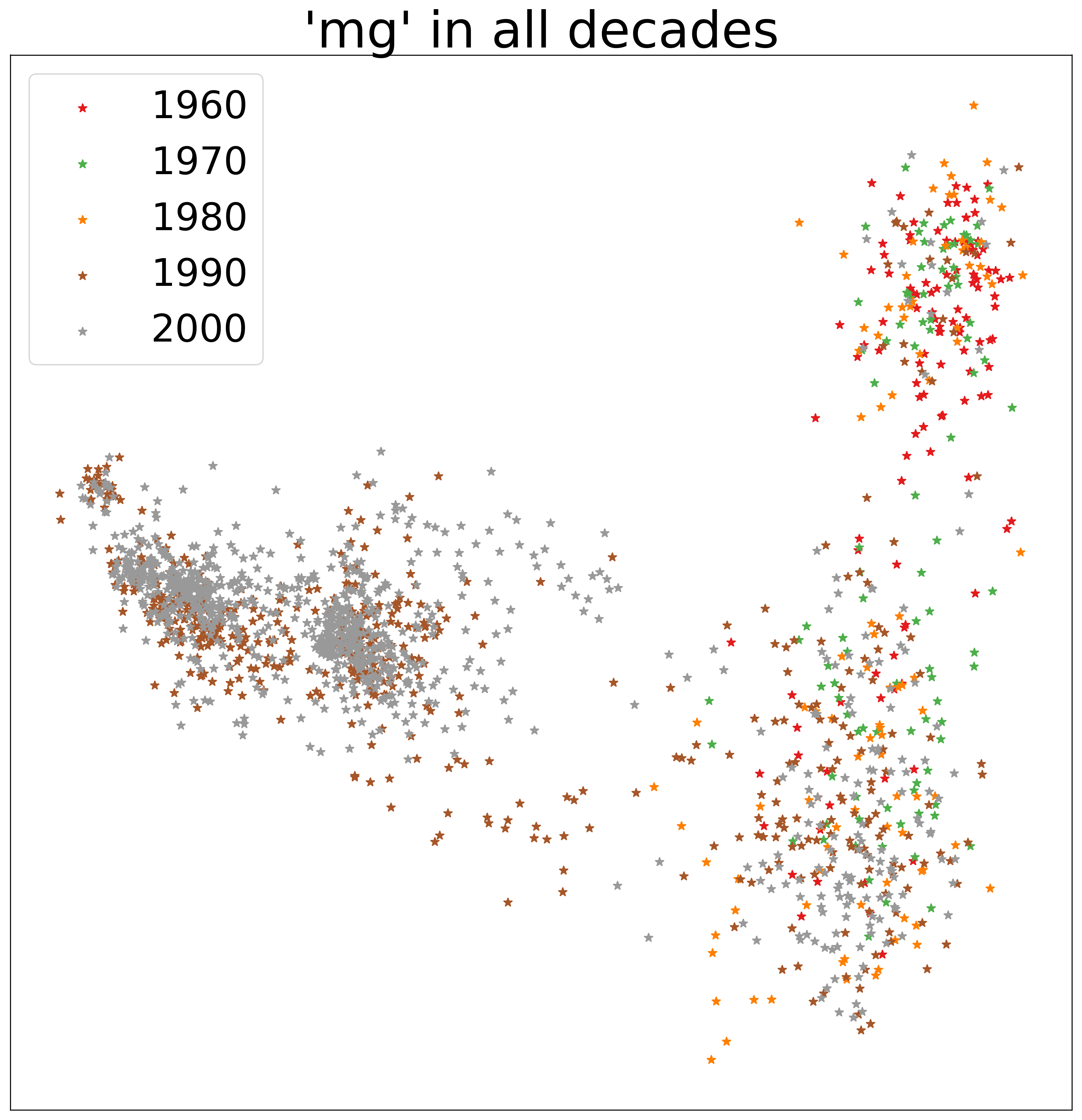}
    \includegraphics[width=0.48\linewidth]{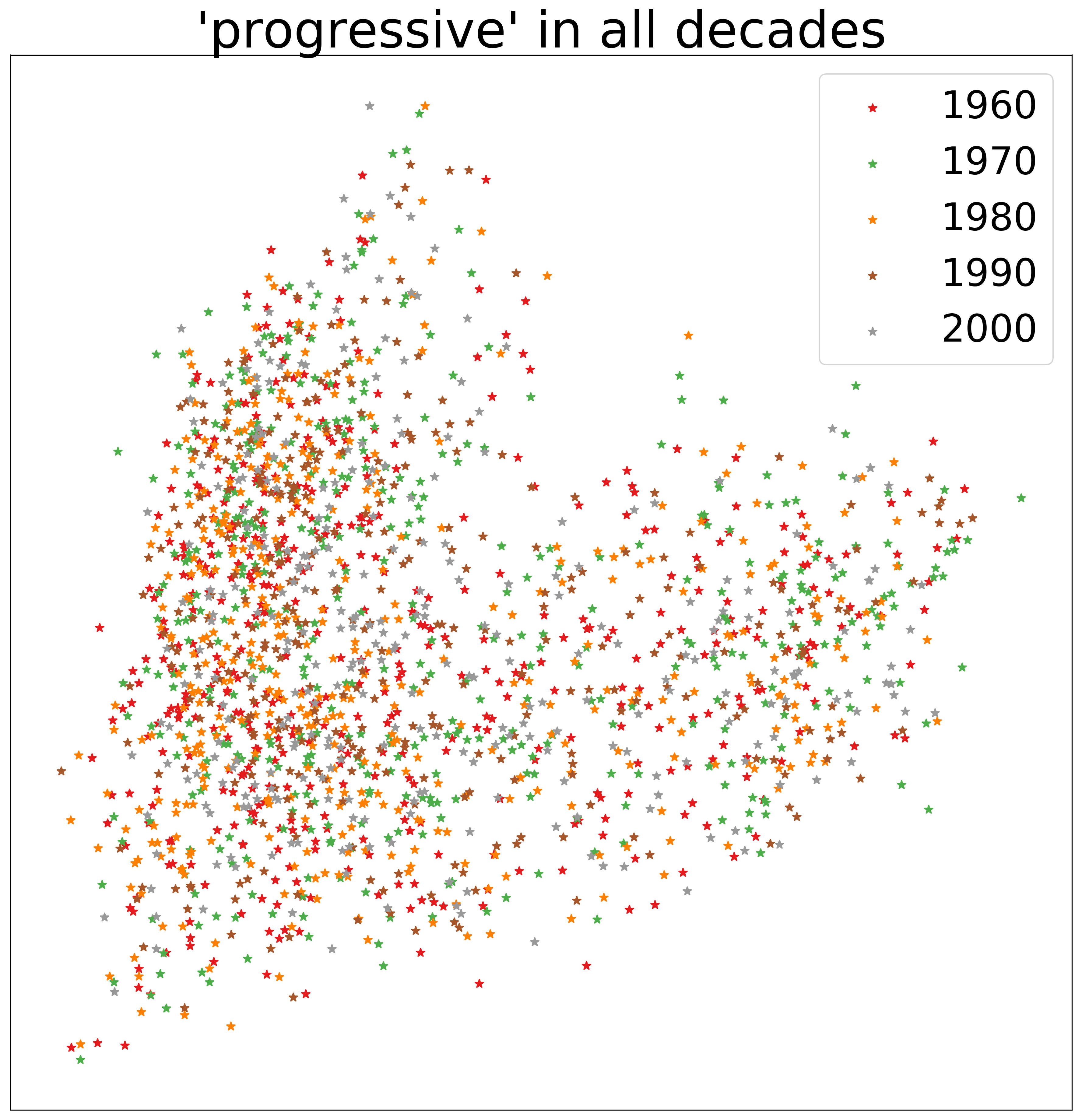}
    \includegraphics[width=0.48\linewidth]{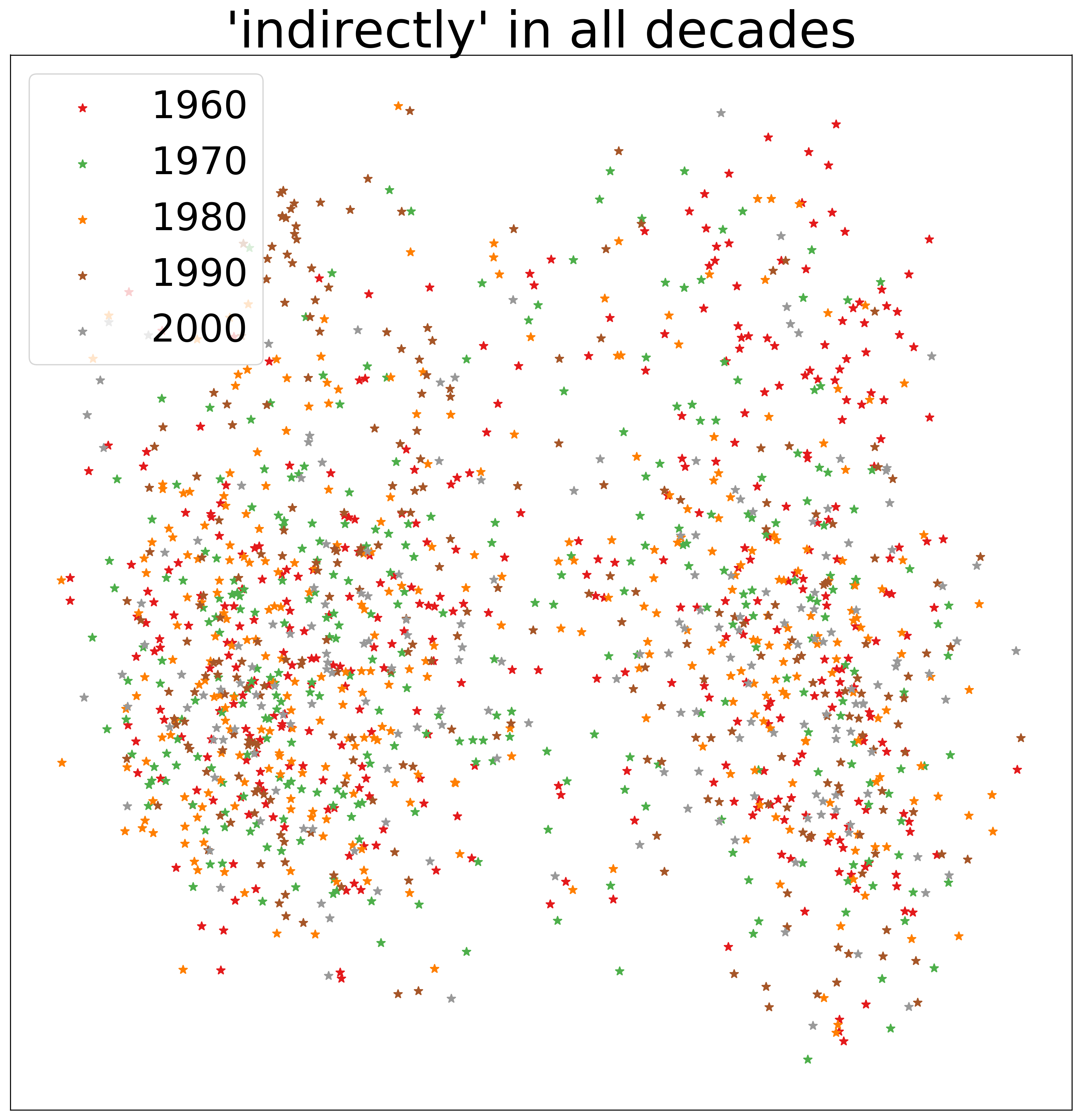}
    \caption{PCA projections of token embeddings for \word{banish}, \word{mg}, \word{progressive} and \word{indirectly} across all 5 COHA decades.}
    \label{fig:dia_strange_cases}
\end{figure}

Figure~\ref{fig:dia_strange_cases} shows the PCA projections of token embeddings for four of the words from Table~\ref{tab:points_change} across the five COHA decades. Below we describe these diachronic vector spaces  more closely to explain the nature of each category of `problematic' words.

\word{Progressive} (in the bottom left part of the plot) belongs to \textcolor{red}{the 1st class} and presents the easiest case to explain. As can be seen from the plot, the occurrences from all five decades are spread uniformly over the vector space. There are no regions inhabited by occurrences only from some subset of the decades. This means no sense was acquired or lost at any point in time. The reason for the  high absolute value of the change score is the context-dependent meaning of the word itself. Actually, it featured high change scores in all the previous decade pairs as well: 0.781, 0.780, 0.778. Its contexts are so diverse and `fluid' that PRT/APD detects strong change whatever corpora are under comparison. In this respect, \word{progressive}, \word{designate}, \word{form} and similar entries behave much like function words: their contextualized embeddings are in a constant flux. Such cases can be traced and discarded when we have a sequence of several time bins clearly showing the constant character of the changes. However, if looking at one pair of time bins only (like in the SemEval 2020 Task 1), a researcher can be mistaken into concluding that an actual semantic shift is undergoing here.

\word{Indirectly} and \word{mg} (bottom and top right parts of the plot correspondingly) belong to \textcolor{magenta}{the 2nd class} and they do reflect some actual changes in the corpora. The plot for \word{indirectly} features a small cluster of the 1990s occurrences in the top left corner.  Otherwise, the occurrences from different time bins are spread uniformly, so this must be the reason of the detected `change'. Indeed, for this word we find high change coefficients both for the 1990s (0.779) and the 2000s (0.780), while before that the scores were much lower. Accordingly, something had happened to \word{indirectly} in the 1990s and then arguably went back to normal in the 2000s. Manual inspection of the 1990s-specific cluster reveals sentences like those in example~\ref{ex:indirect_1990}:

\begin{covexample}
\begin{enumerate}
    \item `Lane now holds 1,966,692 shares directly and \textbf{indirectly}, worth \$ 17,700,228.'
    \item `Parshall now holds 300 Class A shares \textbf{indirectly}, worth \$ 3,975.'
\end{enumerate}
\label{ex:indirect_1990}
\end{covexample}

All of them are excerpts from a long text titled `Depressed shares are a hit with bargain-hunting execs Banks, utilities among winners', apparently published in the `Insider trading' magazine in 1994. It abounds with reports on various persons holding various amounts of shares directly or indirectly. This type of texts is unusual for COHA: there are no sentences mentioning both \word{hold} and \word{indirectly} simultaneously in other decades, except only one such sentence in the 1980s. Meanwhile, the 1990s sub-corpus has 27 of them (the size of the outlier cluster we see in the plot). The 2000s sub-corpus does not include such texts any more, and thus we observe an equally strong change back when moving from the 1990s to the 2000s. 

For the word \word{mg} (milligram) the situation is similar, except that the change score of 0.792 in the 1990s was the only burst (for other decade pairs, the change scores do not exceed 0.71). It means that something changed in the 1990s and stayed like this through the 2000s. Inspecting Figure~\ref{fig:dia_strange_cases} (top right plot) shows that there is indeed a clearly separated cluster consisting only of the 1990s and 2000s tokens. In the corpus, they always occur in the phrase \word{mg cholesterol}, in sentences like in example~\ref{ex:cholesterol}, being part of dish recipes.

\begin{covexample}
`Per serving: 525 calories, 34 gm protein, ... 674 \textbf{mg cholesterol}, 6 gm saturated fat, 409 mg sodium', 
\label{ex:cholesterol}
\end{covexample}

\word{Cholesterol} did occur in  COHA before the 1990s, but never in such a context (123 occurrences of \word{mg cholesterol} in the 2000s, 128 in the 1990s, and 0 before that).  

In these cases, no semantic shifts in the mainstream sense of this term occurred: the word \word{indirectly} still had the same general meaning in the 1990s, and the word \word{mg} in the 1990s and 2000s. However, the PRT/APD method indeed detected anomalous contextual variances in the corpora under analysis. Another interesting case belonging to this type is the word \word{neutral}, also appearing in Table~\ref{tab:points_change}. Its 2000s burst is caused by the emergence of the frequent collocation \word{gender neutral}, which is missing (or extremely rare) in the previous decades. Are we observing a new sense gradually appearing, or is it just contextual fluctuation? Anyway, independent of whether these variances are due to real changes in the word usage (caused by social and cultural developments) or due to improper corpus collection procedure, they are still really existing bursts in the data. In this respect, this type of controversial predicted changes is different from \word{progressive} or \word{designate}. This is another manifestation of a larger NLP problem of domain sensitivity \cite{okurowski-1993-domain}. Essentially, what the model detected was a domain change in comparison to overall genre structure of COHA.

Finally, the word \word{Banish} belongs to the \textcolor{blue}{proper names subset of the 2nd class}. It features a clearly separated cluster of token embeddings containing exclusively the 1990s occurrences (bottom of the plot). All of them are mentions of `Banish' as the name of one of the characters of the 1996 novel `The Standoff' by Chuck Hogan, see example \ref{ex:banish_1990}:

\begin{covexample}
\begin{enumerate}
    \item `\textbf{Banish} slipped deeper into thought.'
    \item `\textbf{Banish} smiled weakly at the sentiment.'
    \item `The sound man eyed him as he stepped inside, saying nothing about \textbf{Banish's} burnt face.'
\end{enumerate}
\label{ex:banish_1990}
\end{covexample}

The novel is included in COHA almost in its entirety, obviously bringing in a lot of \word{banish} usages very different from its mainstream verbal meaning (recall that we both lemmatize and lower-case our texts). This leads to the high change coefficients in the 1980--1990 pair: 0.794, a strong burst compared to 0.733 (1960s--1970s) and 0.730 (1970s--1980s). Note that the change score is high again when looking at the 1990--2000 pair (0.793). The obvious reason is that the 2000s corpus does not mention Banish from `The Standoff' at all, so the meaning of \word{banish} has returned to its pre-1990s state (more or less equally distributed between the senses of \sense{to expel} and \sense{to destroy, to end}).

Using \word{Banish} in this way is certainly creative, and even more importantly, these occurrences indeed denote something different from the regular meaning of \word{banish}. It can be disputed whether using a verb (or a common noun) as a proper name \textit{is} coining a new sense. Note, however, that a very similar case of the word \word{apple} acquiring the new sense of a well-known company proper name is often used as a classic example for word sense disambiguation \cite{manion2014unsupervised}. From this point of view, \word{banish} certainly temporarily acquired a new sense in the COHA 1990s corpus, and thus the predicted change score perfectly reflects the reality. On the other hand, one could argue that this is true for the title-cased \word{Banish} only, but yielding high change score for \word{banish} is an error. See more on that in subsection~\ref{subsec:remedies}.

\begin{figure}
    \centering
    \includegraphics[width=0.45\linewidth]{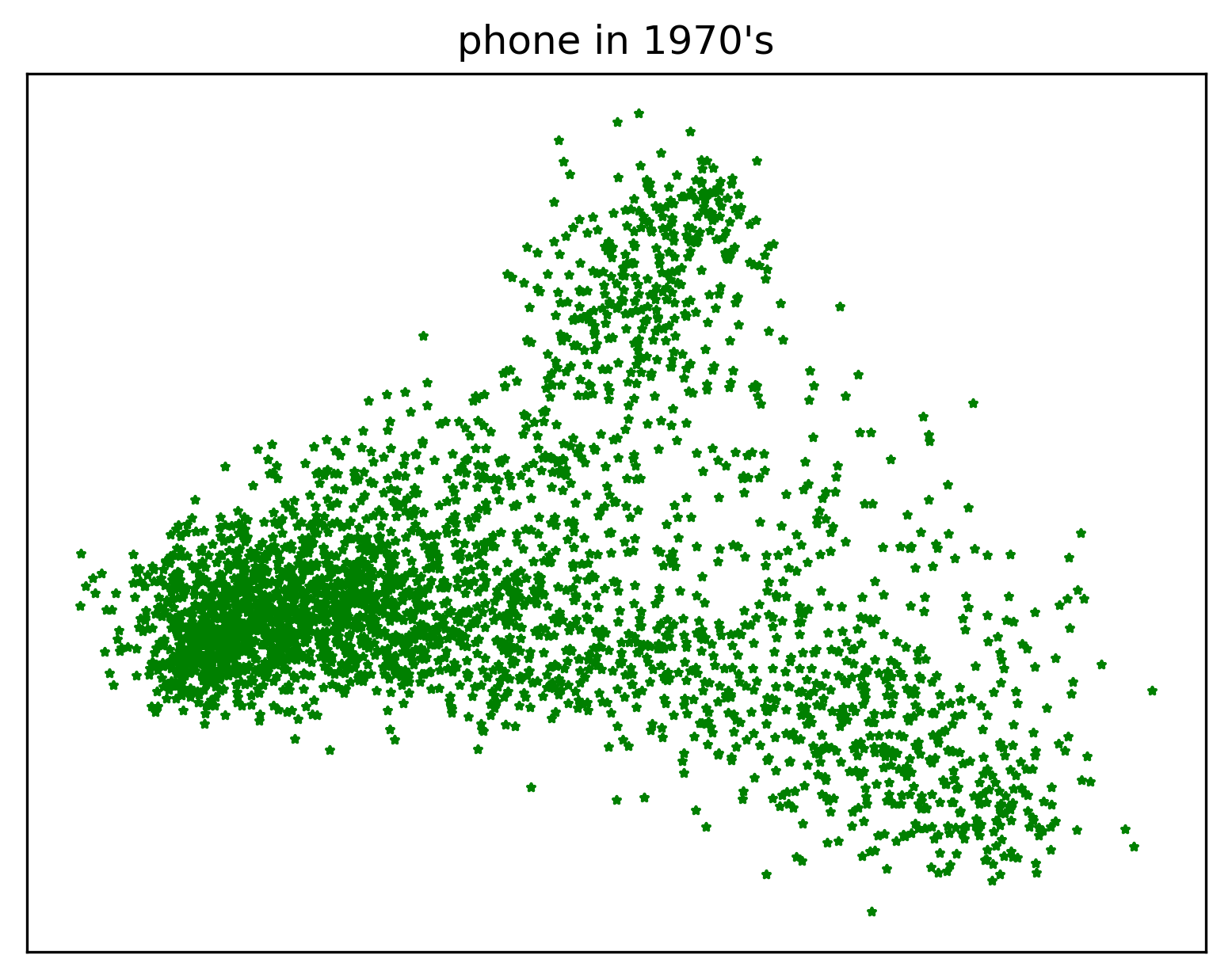}
    \includegraphics[width=0.45\linewidth]{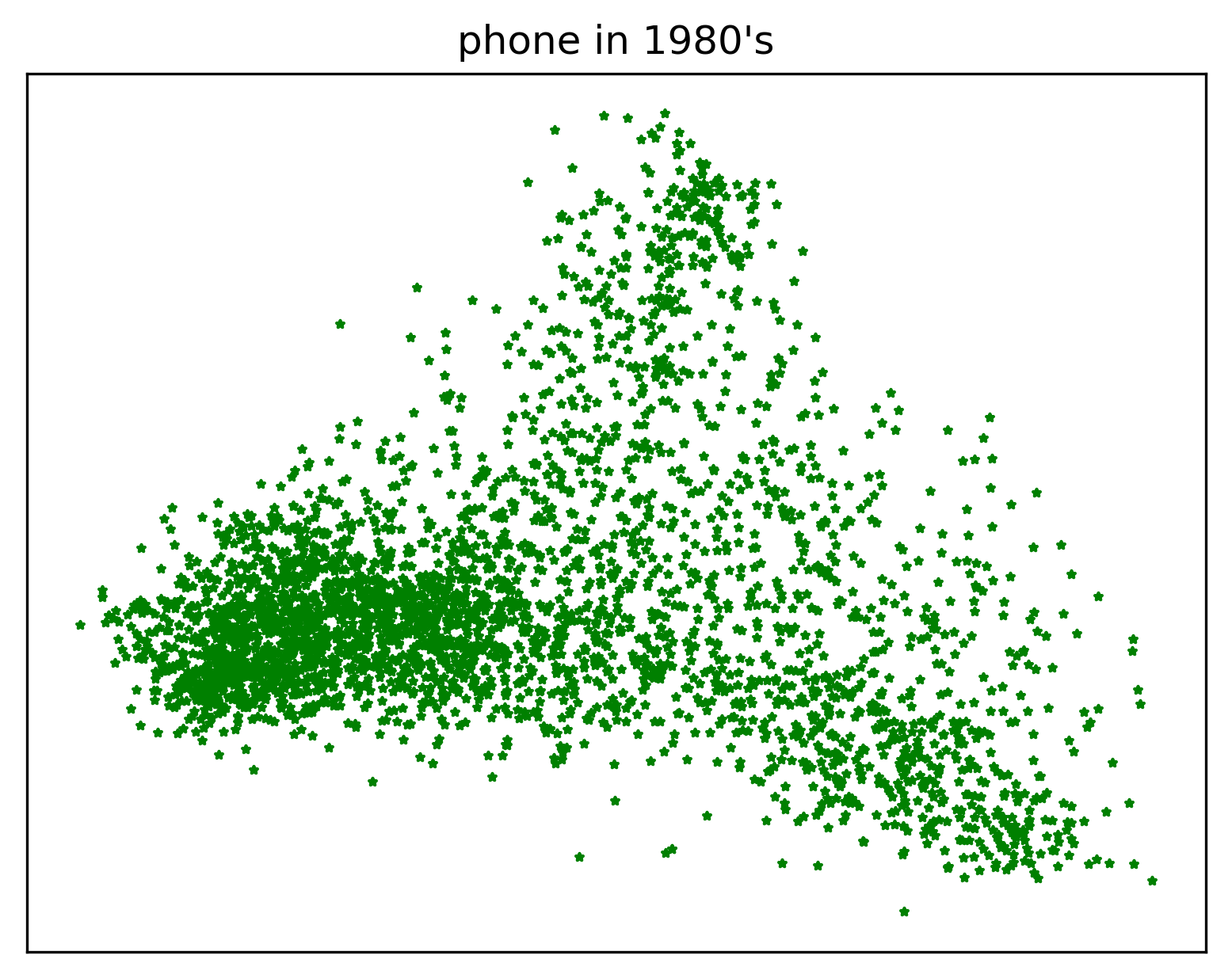}
    \includegraphics[width=0.45\linewidth]{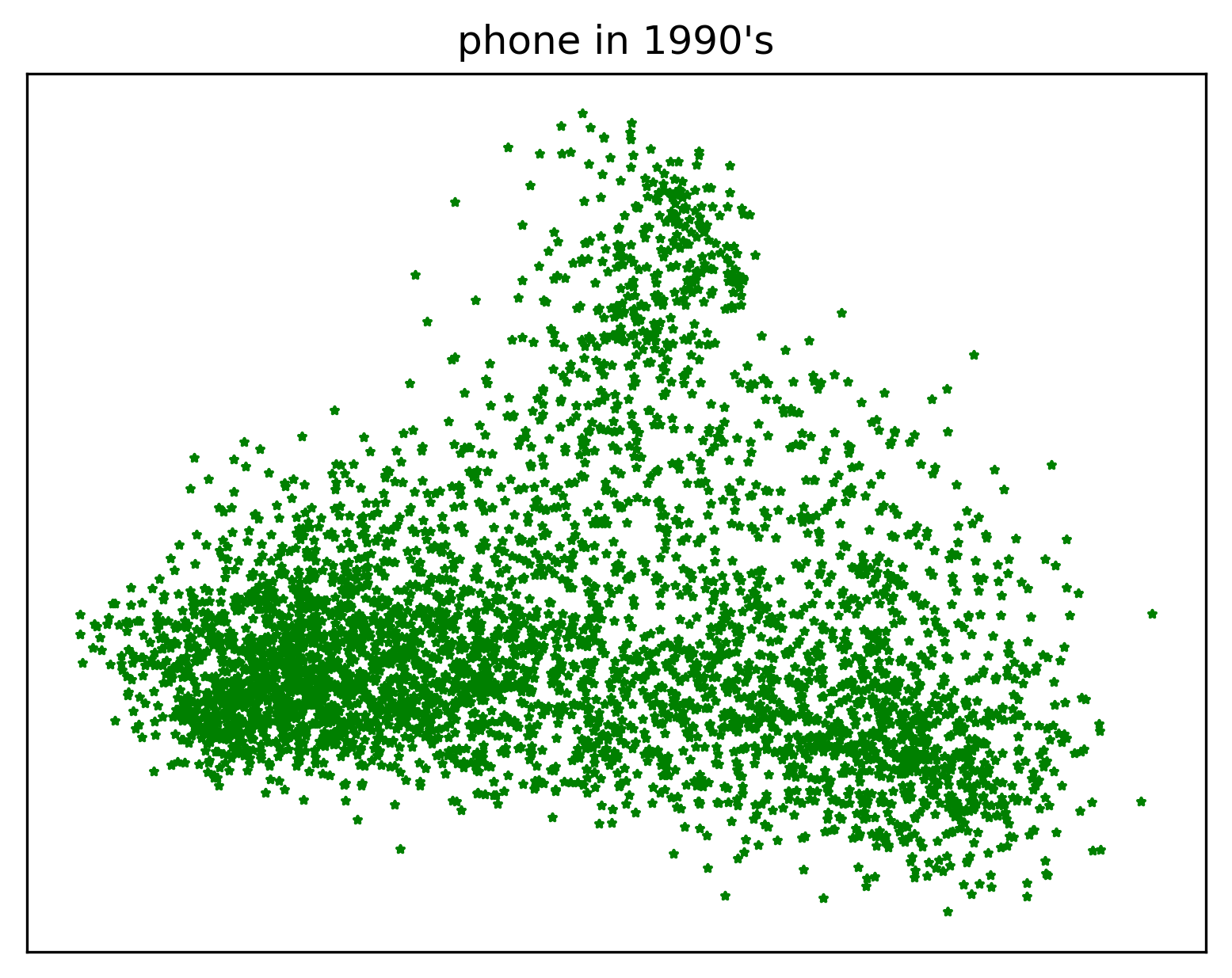}
    \includegraphics[width=0.45\linewidth]{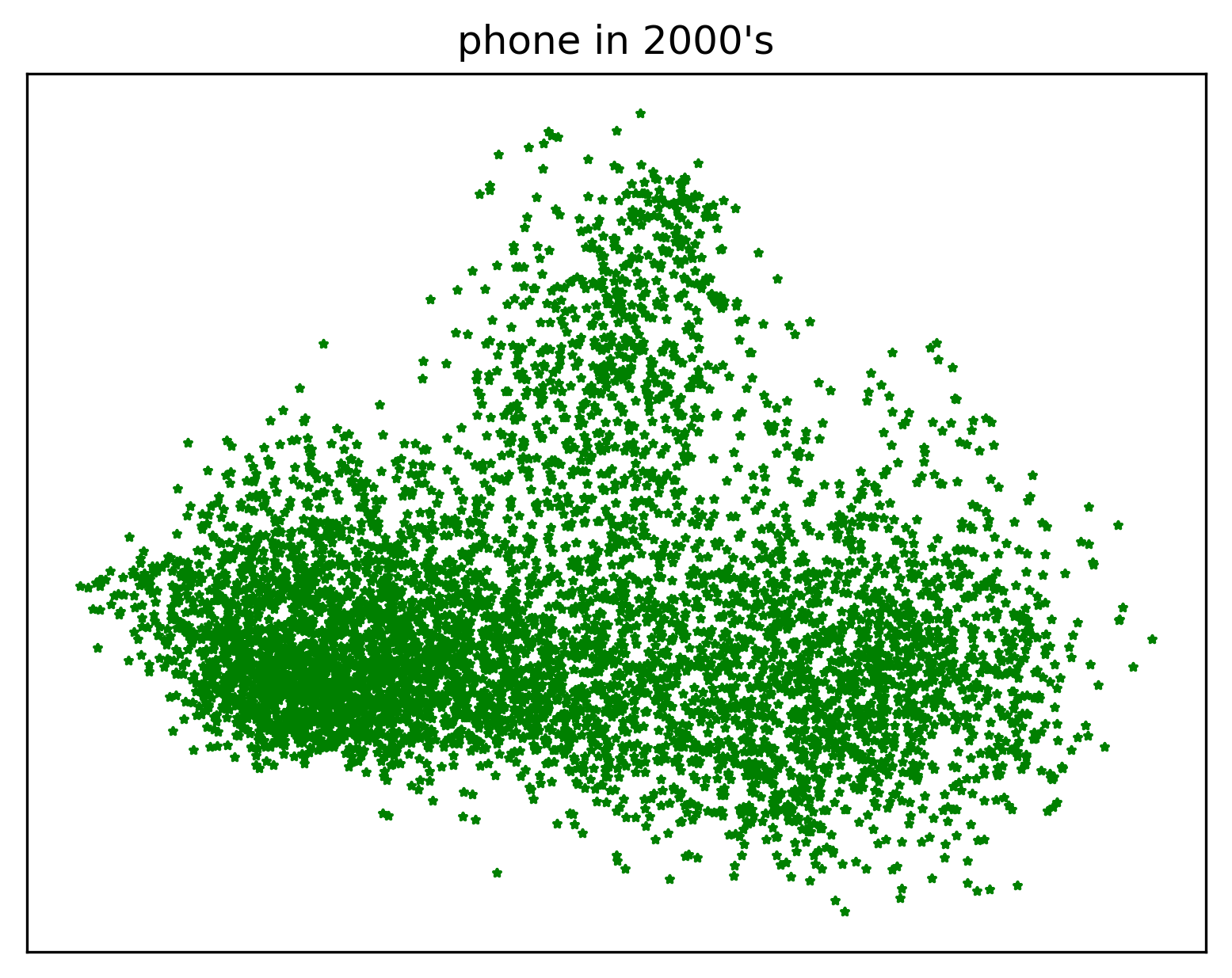}
    \caption{PCA projections of token embeddings for \word{phone} in four different decades: stable syntactic clusters.}
    \label{fig:phone_dia}
\end{figure}

During our manual analysis (following the same workflow of randomly sampling and examining about 20 usages from the core area of  the cluster) we also observed multiple cases where token embedding clusters of an unambiguous word manifested this \textcolor{brown}{word being used in different syntactic roles}. 
For example, the word \word{phone} features three clusters of token embeddings, stable across time (Figure~\ref{fig:phone_dia}). They group occurrences not on \textit{semantic}, but more on \textit{syntactic} grounds:
\begin{enumerate}
    \item \word{phone} is a subject: `Then the \textbf{phone} rang.' (the top cluster)
    \item \word{phone} is an object or an oblique argument: `...took a deep breath and grabbed the \textbf{phone}.' (the bottom left cluster)
    \item \word{phone} is a modifier part of a compound noun: `Please include a daytime \textbf{phone} number.' (the bottom right cluster)
\end{enumerate}

This constitutes \textcolor{brown}{the 3rd class of problematic change predictions}. If the syntactic role frequency distribution of a particular word changes diachronically, the change detection methods based on contextualized embeddings would be triggered by this. As a result, a syntactic shift will be taken for a semantic one. \word{Traditionally} from Table~\ref{tab:points_change} is such an example: for some reason, the 1990s COHA sub-corpus contains much fewer usages of this word as an adjective modifier (\word{traditionally christian}, \word{traditionally male}, etc) than the other decades. 
Interestingly, this syntactic influence is expressed even though we extracted representations from the \textit{top layer} of ELMo, which was shown by \newcite{peters-etal-2018-dissecting} to mostly contain \textit{semantic} information. We discuss the possible smarter ways to employ the model layers in the subsection~\ref{subsec:remedies} below.

\subsection{What about static embeddings?}
It can be argued that the issues mentioned above are not specific for contextualized architectures. To test this, we trained five static embedding models on five COHA sub-corpora each representing one of the decades (1960, 1970, 1980, 1990, 2000). We employed the widely used skip-gram with negative sampling (SGNS) algorithm from \newcite{Mikolov_representation:2013}, also known as \textit{word2vec}. The training hyperparameters were set as follows: symmetric context window of 10 words to the right and 10 words to the left, minimal word frequency 5, vector size 300, 10 iterations over the corpus. Then we followed the standard semantic change detection workflow (so called `SGNS+OP') :
\begin{enumerate}
    \item Vector matrices of each model were aligned to the 2000s matrix with the Orthogonal Procrustes (OP) transformation \cite{hamilton-etal-2016-diachronic}; the 2000s decade was chosen as the basis for alignment, since this model has the largest vocabulary (65 246 words).
    \item For each target word, the cosine distances between its aligned static embeddings in the four consecutive pairs of the COHA decades were calculated. This resulted in the $\mathbf{M}_{static} \in \mathbb{R}^{690\times4}$ matrix, analogous to the $\mathbf{M}$ matrix for ELMo embeddings. The values in $\mathbf{M}_{static}$ are change scores inferred from the \textit{word2vec} models.
\end{enumerate}

\begin{table}
    \centering
    \begin{tabular}{l|c|c|c}
    \textbf{Word} & \textbf{Decade pair} & \textbf{Change} & \textbf{z-score}\\
    \midrule
    \word{drew} & 1960s--1970s & 0.892 & 4.19 \\ 
    \word{banish} & 1980s--1990s & 0.836 & 3.71 \\ 
    \word{jessica} & 1960s--1970s & 0.823 & 3.59  \\ 
    \word{fanny} & 1960s--1970s & 0.811 & 3.49  \\ 
    \word{clayton} & 1970s--1980s & 0.801 & 3.41  \\ 
    \word{val} & 1970s--1980s & 0.798 & 3.39 \\ 
    \word{chris} & 1960s--1970s & 0.790 & 3.32  \\ 
    \word{max} & 1980s--1990s & 0.760 & 3.07  \\ 
    \word{joel} & 1980s--1990s & 0.758 & 3.04  \\ 
    \word{josh} & 1980s--1990s & 0.743  & 2.92 \\ 
    \bottomrule
    \end{tabular}
    \caption{10 points of the strongest change in 5 decades of COHA (as measured with static word embeddings).}
    \label{tab:points_change_static}
\end{table}

Top ten change scores in $\mathbf{M}_{static}$ are shown in Table~\ref{tab:points_change_static}. Again, none of these words looks like an example of a genuine semantic shift, although their z-scores are even higher than those in Table~\ref{tab:points_change}. The important thing is that we observe only two words which also appeared at the top of $\mathbf{M}$: \word{banish} (PRT/APD and PRT) and \word{clayton} (PRT). Since static architectures do not yield token embeddings, one cannot analyze the underlying reasons for high change scores, as we did in the previous subsection. However, it is obvious that most (if not all) words at the top of $\mathbf{M}_{static}$ are proper nouns, which is fully in line with the findings in \cite{shoemark-etal-2019-room}. This makes the predictions of the static models a bit more similar to those produced with the PRT method (which makes sense, since both PRT and static embeddings `merge' all occurrences of a word into a single vector representation), but still substantially different from what any tested contextualized approach yields.

To some extent, the SGNS-OP predictions are potentially easier for `de-noising': one simply has to filter out proper names, which is technically straightforward. Anyway, the take-away message here is that the majority of the problematic examples' categories we mentioned above indeed seem to be specific to contextualized architectures and not manifested in approaches based on static embeddings (which can have their own issues, of course).   

\subsection{Summarizing reflections} \label{subsec:reflections}

Although contextualized architectures are indeed promising for the tracing of diachronic semantic change (especially for finding supporting examples from the corpus), their usage is not entirely straightforward. When measuring the strength of lexical semantic change with contextualized embeddings, one should watch out for the three classes (and one sub-class) of possible unexpected results described above.  A word occurrence can receive a very different token embedding not because the word has acquired a new sense, but because it is used in an unusual syntactic role, or because it is surrounded by unusual neighbors (for example, when the domain of the underlying texts has changed). Since the resulting semantic change score is a derivative of the arrays of token embeddings, one observes strong bursts which manifest changes in contextual variance of a word, not a semantic shift in the lexicographic meaning of this term. 
This is probably not what a historical linguist expects to see, although it can depend on the particular study and the working definition of `semantic shift'.

Note that the problems described here are not entirely novel and have been discussed before in semantic change literature. They are also related to complicated questions about the nature of meaning and of what exactly it means to undergo a `semantic shift', especially when we observe a case of contextual variance. If we stick to the distributional view that `senses are in fact clusters of corpus usages' \cite{kilgarriff1997don}, the cases described above should definitely count as sense inventory changes, or at least the appearance of short-term senses which then fade away. If one does not employ external data sources (like ontologies or diachronic dictionaries), there is no reliable way to discern `semantic changes' from `differences in the underlying textual data': they are simply the same thing. 

All this is an inevitable consequence of accepting the data-driven distributional paradigm. It can be argued that any distributional corpus-based model suffers from these problems by definition, simply because it derives its signal from contexts surrounding word tokens. In fact, the `clusters' on the plots in this section can be more properly described not as `senses', but as `sense nodules' (`lumps of meaning with greater stability under contextual changes') from \newcite{cruse2000aspects}.
However, it is now confirmed that this fundamental issue is still present in deep contextualized language models, often thought to be superior to their static type-based predecessors. Addressing it is a challenge facing the semantic change detection community in general. Before this issue is solved, the output of current semantic change detection models still needs human scrutiny, unless the downstream task at hand is tolerant to high amounts of false positives.

\subsection{Possible remedies} \label{subsec:remedies}
This paper is aimed rather at results interpretation and analysis than at improving task scores. With this in mind, we here do not offer fully implemented and evaluated solutions addressing the issues described above. Still, in this subsection, some possible thought directions are outlined (they are by no means exhaustive).

The 1st class (words with `fluid' meaning) is clearly erroneous. These words always exhibit strong change without it being of any significant linguistic interest, and ways must be devised to filter out these cases. Possible approaches to do this could include measuring change scores between random subsets of the same time bin: if they are as high as those between different time bins, the possible reason is the word's fluidity, not real semantic change.  

The 2nd class (`data bursts') can be considered erroneous or not, depending on one's definition of semantic change (e.g., whether it includes contextual variance). It can be looked at as a corpus problem: COHA is not entirely well-balanced with respect to sense distribution. 
On the other hand, any dataset is biased and incomplete, and the notion of a `100\% balanced' corpus is in fact ill-defined (balanced \textit{for what}?). Arguably, the creators of COHA did not set an aim to somehow `properly represent' the distribution of word senses (even if there existed robust methods to implement this). As \newcite{hengchen2021challenges} put it, `whatever is encountered in corpora is only valid for those corpora and not for language in general'.
For the subclass of proper names, pre-processing decisions can help: keeping proper names capitalized will avoid them mixing with common nouns and predicting a shift for an otherwise stable noun which just happens to have a popular proper name counterpart. On the other hand, this raises difficult questions about the boundaries between word types and about the correctness of separating `Apple' from `apple' based on their written forms only. Again, what constitutes an error here has to be decided separately for each particular study.

To detect the cases belonging to the 3rd class (syntactic shifts), one can arguably use the distributions of PoS tags surrounding a given word. However, this approach is not scalable except for the cases when we are interested in a small closed set of target words only. Another option is learning a weighted function of different layers of the language model (both lower layers carrying more syntactic information and higher layers carrying more semantic information) to properly discern between changes on different language tiers.

In any case, this will require a human annotated dataset of changes of different types. With this at hand, it will be possible to train a meta classifier taking as an input the PRT and APD change coefficients (including signals from different network layers), frequency values, capitalization and other features mentioned above and producing a binary decision on whether the current data point is potentially a false positive. 

\section{Limitations}
Our analysis in Section~\ref{sec:shift_bad} was based on the top $10$ most changed words according to each change detection method. We acknowledge that more insights can be obtained by analyzing more top ranking words (this is also true for static embeddings).

Another important limitation of this work is our focus on false positives: that is, words which are assigned a high semantic change score when this arguably should not be the case.  The study of false negatives (words known to have changed but assigned low scores by the models) is a topic of its own.  It is related to possible analysis of the PRT, APD and PRT/APD predictions on the `stable' versus `changed' words from the SemEval-2020 test set \cite{schlechtweg2020semeval}. We hope to deal with these aspects in the future.

The plots in Sections~\ref{sec:shift_good} and \ref{sec:shift_bad} show token representations of our target word. A potentially more powerful visualization approach could include showing also some `anchor' or `seed' words serving to better disambiguate senses of different tokens (or time-dependent representations for static word embeddings). Note, however, that choosing such anchor words is a separate task in itself, see, for example, \citet{hamilton-etal-2016-diachronic}.
In addition, the plots could arguably be made more visually enticing and insightful by using different markers and sub-sampling of data points (to make the plots look cleaner). This was out of scope for this work.

\section{Conclusion}

We have qualitatively analyzed the outputs of contextualized embedding-based methods for detecting diachronic semantic change. First, we improved the  results of prior work by proposing an ensemble of two methods from \newcite{kutuzov2020uio}, which proved to be a robust solution across the board, outperforming prior contextualized methods on the SemEval-2020 Task 1  test sets \cite{schlechtweg2020semeval} and on the GEMS test set \cite{baroni:2011}. 
Our `PRT/APD' method is more suitable for a realistic case of not knowing the gold score distribution beforehand.

Using PRT/APD together with ELMo, we produced semantic change coefficients for 690 English words across five decades of the 20 and 21 century using the COHA corpus \cite{coha}, and systematically examined these predictions. Although many cases of strong detected change do correspond to well-known semantic shifts, we also found multiple less clear-cut cases. These are the words for which a high change score is produced by the model, but it is not related to any `proper' diachronic semantic shift (not causing a new entry in a dictionary). We discuss such cases in detail with examples, and propose their linguistic categorization. Note that these issues do not depend on a particular training algorithm (or an ensemble of algorithms). There is no reason for them to not appear also when using  BERT or any other token-based embedding architecture; see \newcite{giulianelli-etal-2020-analysing} and \newcite{yenicelik-etal-2020-bert} who show that BERT generates representations which form structures tightly coupled with syntax and even sentiment. To properly test it empirically could be an interesting future work, but we have already shown that semantic change detection approaches based on static word embeddings (as opposed to contextualized token-based architectures) yield different sorts of problematic predictions.  

It is not immediately clear whether improving the quality and representativeness of diachronic corpora can help alleviating this issue (producing more historical data is often not feasible if not impossible). Still, it would be interesting to refine our results using larger or cleaner historical corpora: for example, Clean COHA \cite{alatrash-etal-2020-ccoha}. We also plan to analyze the semantic change modeling results for other languages besides English, as well as using different neural network layers to infer semantic change predictions.

The data (change scores for all target words) and code (including visualization tools) used in this work is available at \url{https://github.com/ltgoslo/lscd_lessons}.

\bibliographystyle{nejlt_bib}
\bibliography{nejlt}

\end{document}